\documentclass[11pt, logo, onecolumn, copyright, colorlinks=true, allcolors=blue]{nvidiatechreport}
\usepackage{graphicx} 
\usepackage{subcaption,ragged2e}
\usepackage[round]{natbib}

\usepackage[normalem]{ulem}
\usepackage[utf8]{inputenc} 
\usepackage[T1]{fontenc}    
\usepackage{url}
\usepackage{tikz}
\usepackage{enumitem}
\usetikzlibrary{positioning}
\usepackage{array}
\usepackage{booktabs}
\usepackage{wrapfig}
\usepackage{caption} 
\usepackage{listings}
\usepackage{adjustbox}
\usepackage{footnote}
\usepackage{multirow}
\usepackage{amsmath}
\usepackage{amsfonts}
\usepackage{breakcites}
\usepackage{blindtext}
\usepackage{svg}
\usepackage[most]{tcolorbox}
\usepackage{tabularx}

\newcommand{\newparagraph}[1]{\noindent\textbf{#1\hspace{0.5em}}}

\title{Nemotron-H: A Family of Accurate and Efficient Hybrid Mamba-Transformer Models}
\author{\large NVIDIA}
\date{}

\begin{document}

\begin{abstract}
\large \textbf{Abstract.}
\normalsize
As inference-time scaling becomes critical for enhanced reasoning capabilities, it is increasingly becoming important to build models that are efficient to infer. We introduce Nemotron-H, a family of 8B and 56B/47B hybrid Mamba-Transformer models
designed to reduce inference cost for a given accuracy level. To achieve this goal, we replace the majority of self-attention layers in the common Transformer model architecture with Mamba layers that perform constant computation and require constant memory per generated token. We show that Nemotron-H models offer either better or on-par accuracy compared to other similarly-sized state-of-the-art open-sourced Transformer models (e.g., Qwen-2.5-7B/72B and Llama-3.1-8B/70B), while being up to 3$\times$ faster at inference. To further increase inference speed and reduce the memory required at inference time, we created Nemotron-H-47B-Base from the 56B model using a new compression via pruning and distillation technique called MiniPuzzle. Nemotron-H-47B-Base achieves similar accuracy to the 56B model, but is 20\% faster to infer. In addition, we introduce an FP8-based training recipe and show that it can achieve on par results with BF16-based training. This recipe is used to train the 56B model.
We are releasing Nemotron-H base model checkpoints with support in Hugging Face and NeMo.
\end{abstract}

\maketitle

\section{Introduction}

Attention~\citep{vaswani2023attentionneed} traditionally has been the powerhouse of large language models (LLMs). Yet, to generate one token during auto-regressive inference, the self-attention layer must perform computation that scales linearly with the number of tokens seen so far since it models interactions between all pairs of tokens in a sequence (potentially in a causal way); as a result, self-attention layers also have to store state in a KV cache during inference that is linearly proportional to the number of tokens in the sequence~\citep{kwon2023efficientmemorymanagementlarge}. Recent reasoning LLMs, however, exhibit \emph{inference-time scaling}, where generating more tokens at inference time can improve the quality of model responses~\citep{openai_o1,deepseekai2025deepseekr1incentivizingreasoningcapability}. Thus, attention layers can fundamentally limit overall model intelligence.

To address this issue, much recent work has proposed alternative architectures~\citep{katharopoulos2020transformersrnnsfastautoregressive, beltagy2020longformerlongdocumenttransformer, gu2024mambalineartimesequencemodeling, dao2024transformersssmsgeneralizedmodels}. One such example is a series of \emph{hybrid models} that replace the majority of the self-attention layers in the standard Transformer architecture with more efficient layers (e.g., Mamba, Mamba-2, or sliding-window attention layers) that have sub-linear or even constant compute and memory requirements. Recent work has shown hybrid models to be competitive with more traditional Transformer architectures~\citep{waleffe2024empiricalstudymambabasedlanguage, lieber2024jambahybridtransformermambalanguage, gemmateam2025gemma3technicalreport}.

\begin{figure}[t]
    \centering
    \includegraphics[width=0.93\linewidth]{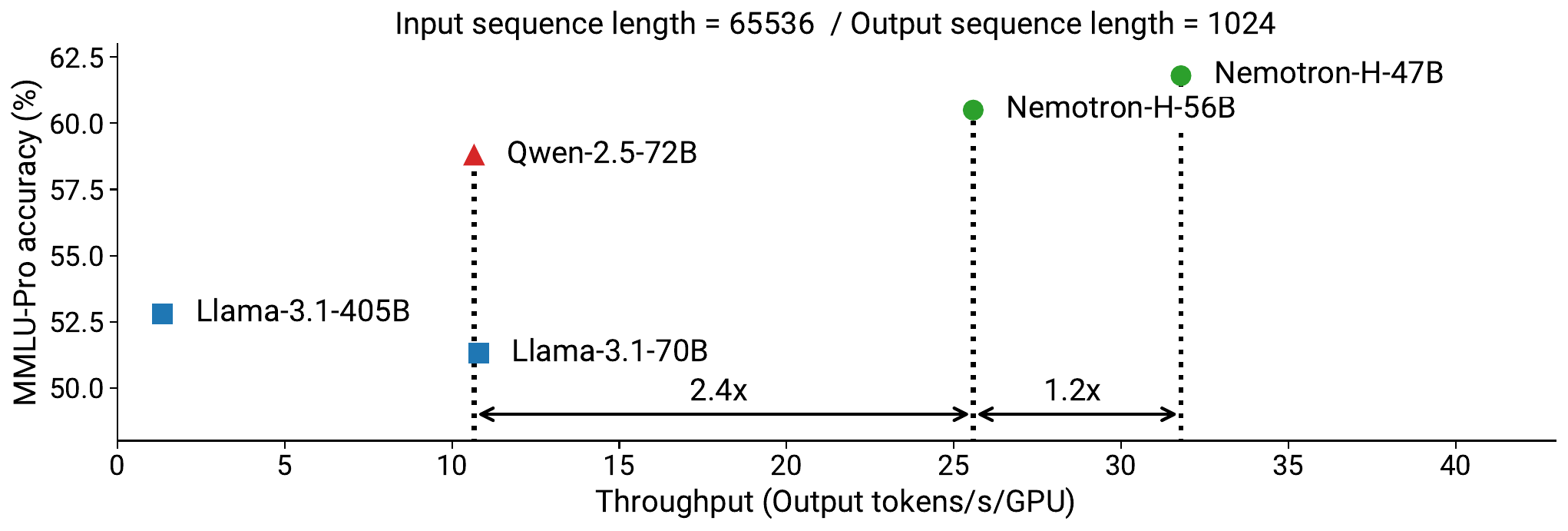}
    \caption{MMLU-Pro accuracy~\citep{wang2024mmluprorobustchallengingmultitask} versus per-GPU inference throughput for Nemotron-H-56B/47B-Base compared to existing similarly-sized Transformer models. Nemotron-H models provide state-of-the-art accuracy and inference time speedups.}
    \label{fig:inference_speed_56b}
\end{figure}

Motivated by improving inference efficiency, we introduce the Nemotron-H family of hybrid Mamba-Transformer models. The Nemotron-H models consist of a mixture of Mamba-2, self-attention and MLP layers, and achieve state-of-the-art accuracy and improved inference speed when compared to open-sourced Transformer models of similar size (Figure~\ref{fig:inference_speed_56b}). The Nemotron-H family has a series of 8-billion-parameter models (Nemotron-H-8B-Base, Nemotron-H-8B-Instruct, Nemotron-H-8B-VLM), and a series of 47/56-billion-parameter models (Nemotron-H-47B-Base, Nemotron-H-56B-Base, Nemotron-H-56B-VLM) that offer either better or on-par accuracy compared to Qwen-2.5-7B/Llama-3.1-8B and Qwen-2.5-72B/Llama-3.1-70B, respectively. For example, Nemotron-H-56B-Base outperforms Llama-3.1-70B on 16 out of 17 tasks that we evaluated (\S\ref{subsec:base_model_evals}).
With larger sequences (65536 input sequence length, 1024 output tokens), we measure up to 3$\times$ higher inference throughput on NVIDIA H100 GPUs.

To achieve the above results, we pre-train Nemotron-H models on up to 20 trillion tokens of high-quality curated and synthetically-generated data~\citep{su2024nemotroncctransformingcommoncrawl, akter2024mindmathinformedsynthetic}. Nemotron-H-56B-Base is the first Nemotron model to be fully pre-trained using a FP8-based recipe. Our recipe, called per-tensor current scaling, is coarse grained and involves quantizing entire tensors using a single scaling factor. This factor is chosen to preserve the largest value in the given tensor; any value too small to fit in the desired FP8 format is flushed to zero. In addition, we find it important for training stability and convergence to leave the first and last 4 GEMMs of the model in BF16. In smaller ablations with 8B-parameter models on token horizons up to 15 trillion tokens, we show that FP8 per-tensor scaling can reach equal or better downstream task accuracy compared to BF16.

To efficiently tailor Nemotron-H models for different deployment scenarios, we introduce a new compression via pruning and distillation paradigm, called MiniPuzzle, that combines the simplicity of Minitron~\citep{sreenivas2024llmpruningdistillationpractice} and the versatility of Puzzle~\citep{bercovich2024puzzledistillationbasednasinferenceoptimized}. MiniPuzzle can be used to turn a larger model into a smaller model satisfying specific memory, parameter count, or latency constraints. We use MiniPuzzle to distill Nemotron-H-56B-Base to Nemotron-H-47B-Base, using only 63 billion training tokens and FP8 training. Nemotron-H-47B-Base can be deployed in FP4 precision on a single NVIDIA RTX 5090 GPU with 32GiB of memory (Nemotron-H-56B-Base's model weights would require roughly 29.5GiB alone, limiting maximum context size).

The Nemotron-H base models can also be effectively post-trained to produce models with high accuracies on vision-language, instruction following, and long-context (e.g., RULER) benchmarks.
Vision-Language Models (VLMs) based on both Nemotron-H-8B and Nemotron-H-56B have already been used to develop very strong reasoning models for physical AI as part of the Cosmos-Reason1 project~\citep{nvidia2025cosmosreason1physicalcommonsense}.

Overall, the Nemotron-H family of models demonstrates that hybrid models can be state-of-the-art in terms of capabilities while offering improved inference speed. Moreover, we believe FP8 pre-training and compression techniques like MiniPuzzle make it cheaper and more efficient to create such models.

\newpage
To enable further research, we have released the following base model checkpoints in Hugging Face and NeMo formats on the Hugging Face and NGC model repositories:
\begin{itemize}
    \item \textbf{Nemotron-H-56B-Base.} \href{https://huggingface.co/nvidia/Nemotron-H-56B-Base-8K}{Hugging Face} and \href{https://catalog.ngc.nvidia.com/orgs/nvidia/teams/nemo/models/nemotron-h-56b-base-8k}{NGC}.
    \item \textbf{Nemotron-H-47B-Base.} \href{https://huggingface.co/nvidia/Nemotron-H-47B-Base-8K}{Hugging Face} and \href{https://catalog.ngc.nvidia.com/orgs/nvidia/teams/nemo/models/nemotron-h-47b-base-8k}{NGC}.
    \item \textbf{Nemotron-H-8B-Base.} \href{https://huggingface.co/nvidia/Nemotron-H-8B-Base-8K}{Hugging Face} and \href{https://catalog.ngc.nvidia.com/orgs/nvidia/teams/nemo/models/nemotron-h-8b-base-8k}{NGC}.
\end{itemize}

The rest of this technical report is organized as follows:
\begin{itemize}
    \item In \S\ref{sec:base_pretrain}, we discuss the Nemotron-H model architecture and the pre-training process (including details on the pre-training dataset and FP8 recipe used).
    \item In \S\ref{sec:distillation_full}, we describe the pruning and distillation methods used for model compression.
    \item In \S\ref{sec:vlm}, we introduce the vision-language models based on the Nemotron-H models.
    \item In \S\ref{sec:alignment}, we show that Nemotron-H models can be extended to create competitive instruct and long-context versions.
\end{itemize}
\section{Base Models and Pre-Training}
\label{sec:base_pretrain}
In this section, we discuss the Nemotron-H-8B-Base and Nemotron-H-56B-Base model architectures, as well as key details in the pre-training process used to produce these models. We also compare Nemotron-H-8B/56B-Base with existing open-source state-of-the-art Transformer models on both accuracy (on common benchmark tasks used for base models) and inference speed.

\begin{figure}[t]
\centering
\includegraphics[width=0.9\linewidth]{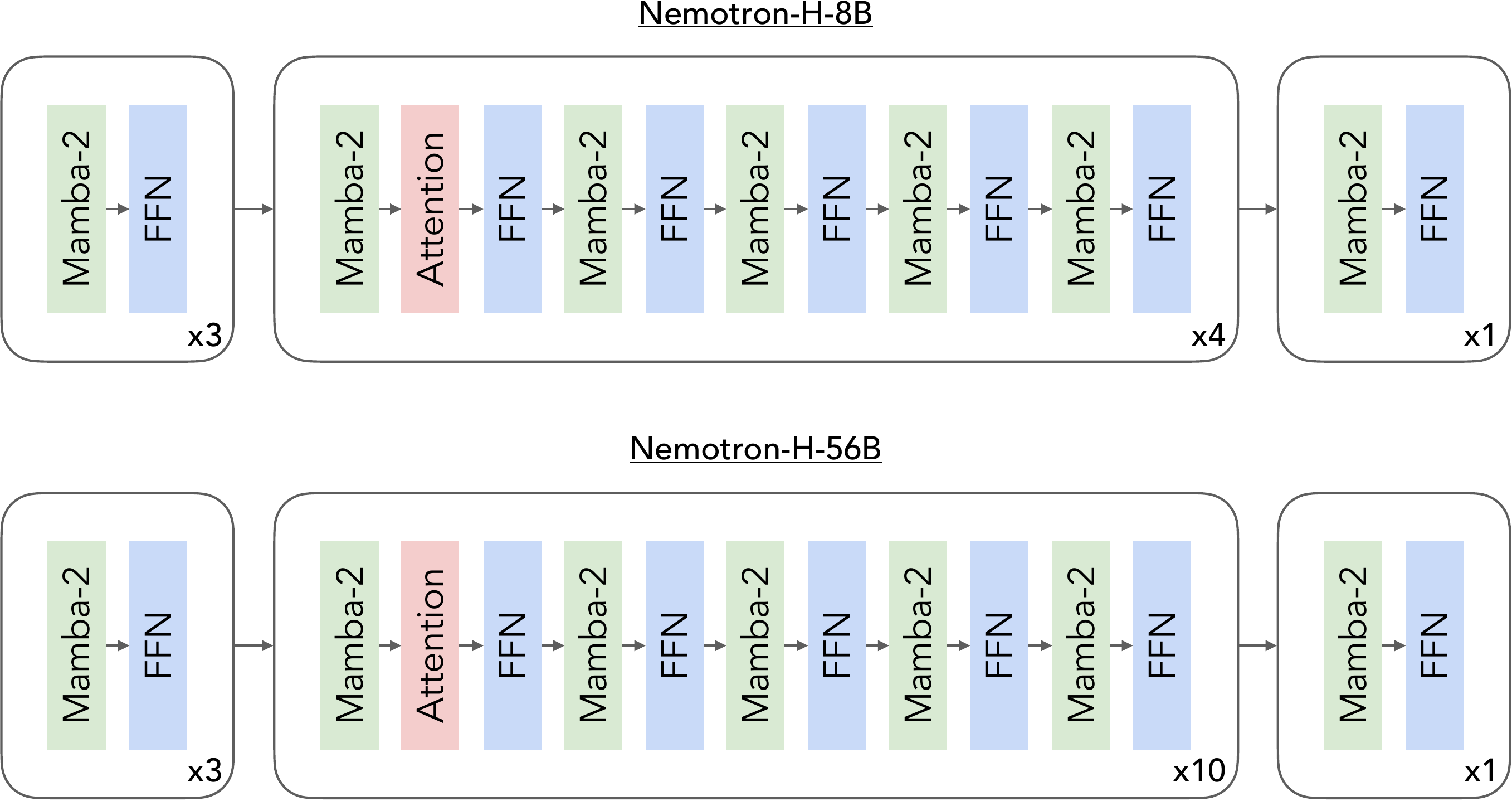}
\caption{Nemotron-H-8B/56B model architectures. Roughly 8\% of the total layers in the model are self-attention layers; these layers are evenly dispersed throughout the model. The rest of the model is made up of alternating Mamba-2 and FFN layers.}
\label{fig:nemotron-h-arch}
\end{figure}

\subsection{Model Architecture}

Nemotron-H models consist of a mixture of Mamba-2~\citep{dao2024transformersssmsgeneralizedmodels}, self-attention, and FFN layers as summarized in Figure~\ref{fig:nemotron-h-arch} and Table~\ref{tab:nemotron-h-arch}. As suggested by prior work~\citep{waleffe2024empiricalstudymambabasedlanguage}, we set the number of attention layers to be roughly 8\% of the total number of layers and evenly disperse them throughout the model. This amounts to 4 self-attention layers (out of 52 layers) for Nemotron-H-8B and 10 for Nemotron-H-56B (out of 118 layers). The rest of the layers consist of an even split between FFN and Mamba-2 layers. We also ensure that a) the first layer in the model is a Mamba-2 layer, b) the last layer in the model is a FFN layer, and c) self-attention layers always precede FFN layers (as they do in a standard Transformer block like in \cite{vaswani2023attentionneed}).

We use a hidden dimension of 4096 for Nemotron-H-8B and 8192 for Nemotron-H-56B. For our smaller model, we use FFN hidden dimension of 21504, 32 attention query heads, and Mamba-2 state dimension of 128; for the larger model, we use FFN hidden dimension of 32768, 64 attention query heads, and Mamba-2 state dimension of 256. Both models use Grouped-Query Attention~\citep{ainslie2023gqatraininggeneralizedmultiquery} with 8 key-value heads, 8 Mamba-2 groups, and squared ReLU~\citep{so2022primersearchingefficienttransformers} activation for FFN layers. We do not use any position embeddings. For Mamba-2 layers, we retain the default values for head dimension (64), expansion factor (2), and window size for convolution (4). We also use RMSNorm~\citep{zhang2019rootmeansquarelayer} for normalization, separate embedding and output layer weights, and no dropout. We do not use bias weights for linear layers. We include a residual skip connection around each Mamba-2, self-attention, and FFN layer in the architecture.

\begin{table}[t]\small
\centering
\begin{tabular}{lcccccccc}
\toprule
        \textbf{Model} &
        \textbf{\begin{tabular}[c]{@{}c@{}}{Number of}\\ layers\end{tabular}} &
        \textbf{\begin{tabular}[c]{@{}c@{}}{Model}\\ dimension\end{tabular}} &
        \textbf{\begin{tabular}[c]{@{}c@{}}FFN \\ dimension\end{tabular}} &
        \textbf{\begin{tabular}[c]{@{}c@{}}Q\\ heads\end{tabular}} &
        \textbf{\begin{tabular}[c]{@{}c@{}}KV \\ heads\end{tabular}} &
        \textbf{\begin{tabular}[c]{@{}c@{}}State \\ dimension\end{tabular}} & 
        \textbf{\begin{tabular}[c]{@{}c@{}}Mamba \\ groups\end{tabular}} \\ \toprule

Nemotron-H-8B & 52 & 4096 & 21504 & 32 & 8 & 128 & 8 \\
Nemotron-H-56B & 118 & 8192 & 32768 & 64 & 8 & 256 & 8 \\
\bottomrule
\end{tabular}
\caption{Summary of the Nemotron-H hybrid Mamba-Transformer architectures.}
\label{tab:nemotron-h-arch}
\end{table}

\paragraph{Nemotron-T-8B Transformer baseline.}
To compare Nemotron-H-8B-Base to a Transformer model in an apples-to-apples fashion, we also trained Nemotron-T-8B-Base on exactly the same data. The Nemotron-T-8B architecture follows the style of GPT-3~\citep{brown2020languagemodelsfewshotlearners}. We use 32 Transformer layers (each has a self-attention layer followed by a FFN layer). As for Nemotron-H-8B, we use a hidden dimension of 4096, 32 query heads, GQA with 8 kev-value heads, a FFN hidden dimension of 21504, squared ReLU activation, RMSNorm, no bias weights for linear layers, no dropout, and separate parameters for model embeddings and output layer weights. We use RoPE for position embeddings~\citep{su2023roformerenhancedtransformerrotary}.

\subsection{Pre-Training Data}
Nemotron-H-8B-Base and Nemotron-H-56B-Base are pre-trained on a large corpus of high-quality curated and synthetically-generated data.

\subsubsection{Curated Data}

We have separate data curation pipelines for four broad data categories: general web crawl data, math data, code data, and ``academic'' data. We discuss each in turn next.

\newparagraph{Web crawl data.} For Nemotron-H, we made several key innovations in our processing of English Common Crawl data compared to Nemotron-4~\citep{parmar2024nemotron415btechnicalreport,nvidia2024nemotron4340btechnicalreport}; these innovations substantially improved data quality. For full details on how this dataset was prepared and various ablations, please refer to the dedicated paper~\citep{su2024nemotroncctransformingcommoncrawl}. We provide a brief summary here.

We focus primarily on extracting as many high quality tokens as possible, so that we can train Nemotron-H for long token horizons (e.g., 15 trillion or more tokens). As is common, we begin with HTML-to-text extraction, language filtering, global fuzzy de-duplication, and exact substring de-duplication. At this stage, however, we deviate from the norm and employ an ensemble of three model-based classifiers to bucket each document into five quality categories. By using an ensemble, we obtain a larger and more diverse set of high-quality tokens compared to approaches that use a single model-based classifier. To retain as many high-quality tokens as possible, we apply heuristic and perplexity filters to the low, medium-low, and medium quality buckets. We also rephrase the low-quality tokens to boost their quality; \S\ref{section:synthetic} describes the synthetic data generation pipeline.

\begin{table}[t]\small
\centering
\begin{tabular}{lcc}
\toprule
\textbf{Quality label} & \textbf{Tokens (billions)} & \textbf{Tokens (\%)} \\ 
\toprule
High & 554 & 8.8 \\
Medium-High & 505 & 8.1 \\
Medium & 2026 & 32.4 \\
Medium-Low & 896 & 14.3 \\
Low & 403 & 6.4 \\
Synthetic-High & 1536 & 24.6 \\
Synthetic-Low & 336 & 5.4 \\ 
\bottomrule
\end{tabular}
\caption{Quality distribution of the 6.3 trillion English Common Crawl tokens.}
\label{table:nemotron-cc-quality-dist}
\end{table}

\begin{figure}[t]
\centering
\includegraphics[width=0.4\linewidth]{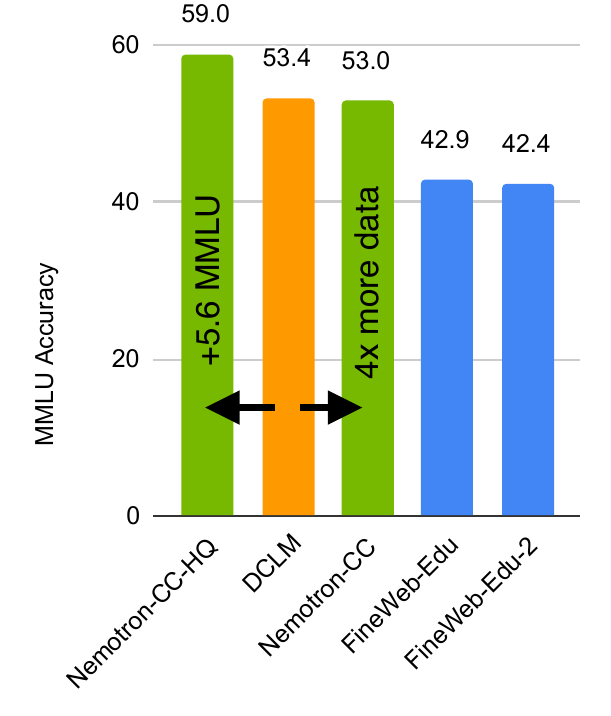}
\caption{MMLU scores for 8B-parameter models trained for 1 trillion tokens. Compared to DCLM, our methods enable us to either create a 4$\times$ larger dataset of similar quality or increase the MMLU using a high quality subset of the tokens. Having a larger dataset (in terms of unique tokens), is crucial when training over long horizons (e.g., 15 or 20 trillion tokens).}
\label{fig:nemotron-cc-figure1}
\end{figure}

The resulting dataset consists of 6.3 trillion tokens, including 4.4 trillion globally de-duplicated ``real'' tokens and 1.9 trillion tokens of rephrased synthetic data. The quality distribution is shown in Table~\ref{table:nemotron-cc-quality-dist}.
As shown in Figure~\ref{fig:nemotron-cc-figure1}, our dataset results in higher-quality models compared to other leading English Common Crawl preparations like DCLM~\citep{li2024datacomp} and FineWeb-Edu~\citep{penedo2024fineweb}: using a high-quality subset increases MMLU by 5.6 points over DCLM, whereas using the full dataset without any weighting of quality buckets achieves comparable accuracy to DCLM while having $4\times$ more unique tokens. Having more unique tokens enables pre-training over longer token horizons without having to do more than 4 to 8 epochs over the full dataset. Higher epoch counts lead to diminishing returns~\citep{muennighoff2024scaling,feng2024maximizedataspotentialenhancing}.

For the benefit of the community, we have publicly released the dataset (as Nemotron-CC\footnote{\url{https://data.commoncrawl.org/contrib/Nemotron/Nemotron-CC/index.html}.}) under the Common Crawl terms of use.

\newparagraph{Math data.} To ensure that technical pages in Common Crawl retain their mathematical content, we leverage the recipe and code base from ~\cite{paster2023openwebmath}. We also find it essential to apply this recipe to other high-quality data sources such as Wikipedia. 
We found that a smaller, higher-quality dataset led to a larger improvement on relevant benchmarks, similar to FineMath~\citep{allal2025smollm2smolgoesbig}. 

\newparagraph{Code data.} We pre-train Nemotron-H models with a considerable number of code tokens. We started from our previous work on Nemotron-4~\citep{parmar2024nemotron415btechnicalreport,nvidia2024nemotron4340btechnicalreport}, and chose to reduce the number of tokens in markup and configuration languages such as HTML, Markdown, YAML, CSS, JSON, and Makefile. Tokens that would have previously come from these languages are instead taken from popular languages like Python, C++, C, and Java.

\newparagraph{Academic data.} 
The Nemotron-H pre-training dataset contains additional tokens gathered from ``high information'' English texts, including permissively-licensed books and articles. We utilize examples across a wide variety of domains (e.g., science, math, humanities, law, and engineering) and a large range of document types including peer-reviewed journal publications, textbooks, published novels, and patent documents.  

As the original data formats of these documents range widely---including EPuB, HTML, XML, plain text, PDF, LaTeX, and markdown---we write custom functions to parse text from the input format into a standardized output format. We maintain appropriate formatting in markdown or LaTeX for complex segments like tables, lists, and equations. We utilize \'Eclair~\citep{karmanov2025eclairextractingcontent} for PDF-to-text extraction. We also develop specialized heuristic filters to remove extraneous information contained within headers or footers of pages. We then apply the Nemotron-4 heuristic and perplexity filters to remove low-quality documents from the set of documents used for pre-training. 

In order to better weight examples from these sources in our final data blend, we developed classifiers to rate documents on their educational content and difficulty. We also try to determine each document's domain (one of biology, business, chemistry, computer science, economics, engineering, history, law, mathematics, medicine, philosophy, physics, psychology, or other). For educational content, we rated all documents on a numerical scale from 0 (no educational information) to 3 (highly relevant educational information). For educational difficulty, we categorized documents with high educational content into the following levels: elementary school, middle school, high school, undergraduate, and graduate. We were able to use these buckets (e.g., ``biology at the undergraduate level with high educational content'') to determine document weights (\S\ref{section:blend}). As expected, we found increasing the weight of highly educational content in important domains at the high school / undergraduate levels to be most helpful.

\subsubsection{Synthetically-Generated Data}
\label{section:synthetic}

We also synthetically generate data to de-noise low-quality data and to increase the diversity of tokens given to the model. We use different processing pipelines to synthetically re-write web crawl, math, and code data. 

\begin{table}[t] \small
    \small
    \centering
    \begin{tabular}{l c l c}
    \toprule
    \textbf{Source} & \textbf{Raw tokens} & \textbf{Prompt} & \textbf{Synthetic tokens} \\
    \toprule
    Low & 403.0 & Wikipedia & 336.3 \\
    \midrule
    \multirow{5}{*}{\textrm{High}} & \multirow{5}{*}{451.3} & Wikipedia & 372.9 \\
                         & & Diverse QA Pairs  & 499.5 \\
                         & & Distill & 157.6 \\
                         & & Extract Knowledge & 303.6 \\
                         & & Knowledge List & 203.2 \\
    \bottomrule
    \end{tabular}
    \caption{Synthetic data token counts (billions) for web crawl, generated with the instruct version of Mistral NeMo 12B.}
    \label{table:nemotron-cc-synthetic-data-stats}
\end{table}

\paragraph{Web crawl data.} We found rephrasing text by LLMs to be an effective way to reduce noise and errors in low-quality crawl data, and produce additional variants of high-quality data with new unique tokens. This leads to better results on downstream tasks.

We rephrased our low-quality data using the ``medium Wikipedia'' prompt from~\cite{maini2024rephrasing}.
For high-quality documents, we generated synthetic data using four additional prompts:
\begin{enumerate}
    \item \textbf{Diverse question-answer (QA) pairs.} Ask questions in various forms (e.g., yes/no question, open-ended question, multi-choice question) about factual information in the text.
    \item \textbf{Distill.} Rewrite the text into a concise and clear passage.
    \item \textbf{Extract knowledge.} Rewrite knowledge from the text and disregard uninformative content.
    \item \textbf{Knowledge list.} Extract key information from the text as an organized list.
\end{enumerate}

These prompts required the model to provide clear and concise responses while preserving factual information and concrete details such as numbers. In total, we generated over 1.8 trillion synthetic tokens using the instruct version of Mistral NeMo 12B. The breakdown of source and generated tokens by quality and prompt is shown in Table~\ref{table:nemotron-cc-synthetic-data-stats}. The full details of this synthetic data generation, including the prompts used, have been published separately in~\cite{su2024nemotroncctransformingcommoncrawl}.

\newparagraph{Math data.}
We use synthetic data to enhance mathematical reasoning benchmarks like GSM8K and MATH~\citep{cobbe2021trainingverifierssolvemath, hendrycks2021measuringmathematicalproblemsolving}, as detailed in~\cite{akter2024mindmathinformedsynthetic}. We expand OpenWebMath~\citep{paster2023openwebmath} from 14 billion to over 100 billion tokens using Nemotron-4-340B~\citep{nvidia2024nemotron4340btechnicalreport}, yielding an 18-point improvement on GSM8K in controlled experiments.

To achieve this, we start with technical pre-training documents from Common Crawl and leverage Nemotron-4-340B to generate dialogues, where a knowledgeable persona guides a less-experienced one (e.g., an interaction between student and teacher). This approach aligns with insights from the Physics of Language Models series~\citep{allenzhu2024physicslanguagemodels32}, and incorporates strategies such as presenting incorrect answers alongside corrections. By structuring content as learning interactions, our method distills broad knowledge from public language models without overfitting to benchmarks. 

\newparagraph{Code data.} With the goal of adding diverse code data centered around problem solving, we chose to synthetically generate mixed natural language and code datasets across 11 programming languages, e.g., Python, C++, C, C\#, and Java. To do this, we prompted Mixtral 8x22B to generate a programming problem inspired by sampled pieces of source code from our curated code dataset. The samples range from 1 to 15 lines of code (LOC), and typically are just a small function. We also prompted Mixtral 8x22B to solve the generated problems and removed clearly invalid solutions on a minimal-effort basis. For example, we extracted the Python code generated by the model and attempted to parse it into an abstract syntax tree (AST), discarding samples which fail to parse. Finally, to form a sample for training, we combined the generated problem and answer into a single sample which is a mix of natural language instruction, generated code (typically enclosed in a markdown code block), and usually an explanation of the approach.

\newparagraph{SFT-style data.} We further add synthetic SFT-style data to the pre-training corpus; this improves the ability of base models to follow instructions. We use the Qwen2.5 series, Mixtral 8x22B, Nemotron-4-340B, and DeepSeek-R1 (only for 56B) models to produce these datasets. We improve math abilities following the pipeline presented in OpenMathInstruct-2~\citep{toshniwal2024openmathinstruct} using carefully curated seed data such as AoPS\footnote{\url{https://artofproblemsolving.com}.}. We also synthesize code data using the approach proposed in Genetic Instruct~\citep{majumdar2024genetic}, with \texttt{tigerbot-leetcode}\footnote{\url{https://huggingface.co/datasets/TigerResearch/tigerbot-kaggle-leetcodesolutions-en-2k}.} and \texttt{the-stack-v2}\footnote{\url{https://huggingface.co/datasets/bigcode/the-stack-v2}.} as seed data. In order to equip the base model with high-quality general knowledge, we generate question-solution pairs with selected topics and domain-specific documents. In total, we add 230 billion synthetic SFT-style tokens (174 billion math tokens, 35 billion code tokens, and 21 billion tokens with general knowledge) to the training corpus.

\begin{figure}[t]
    \centering
    \begin{subfigure}{0.45\textwidth}
        \centering
        \includegraphics[width=\linewidth]{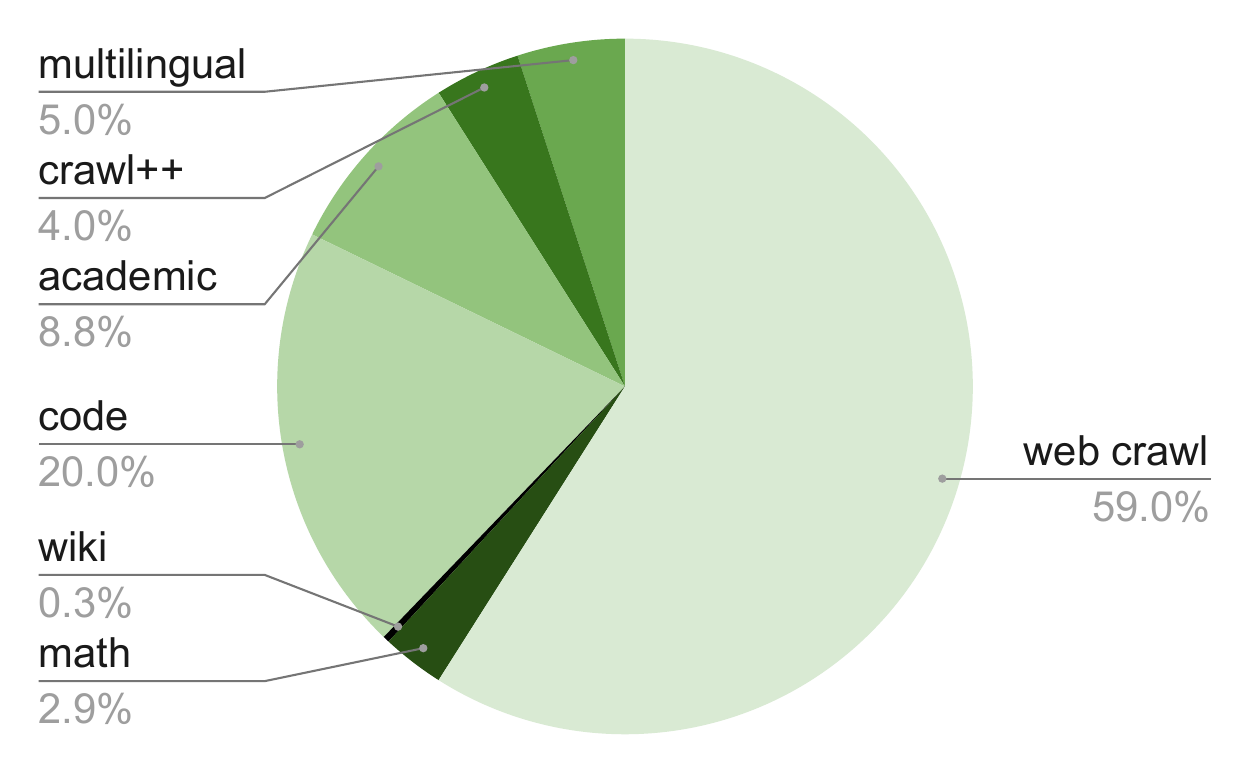}
        \caption{Data mixture of Phase 1.}
        \label{fig:phase1-blend}
    \end{subfigure}
    \hfill
    \begin{subfigure}{0.45\textwidth}
        \centering
        \includegraphics[width=\linewidth]{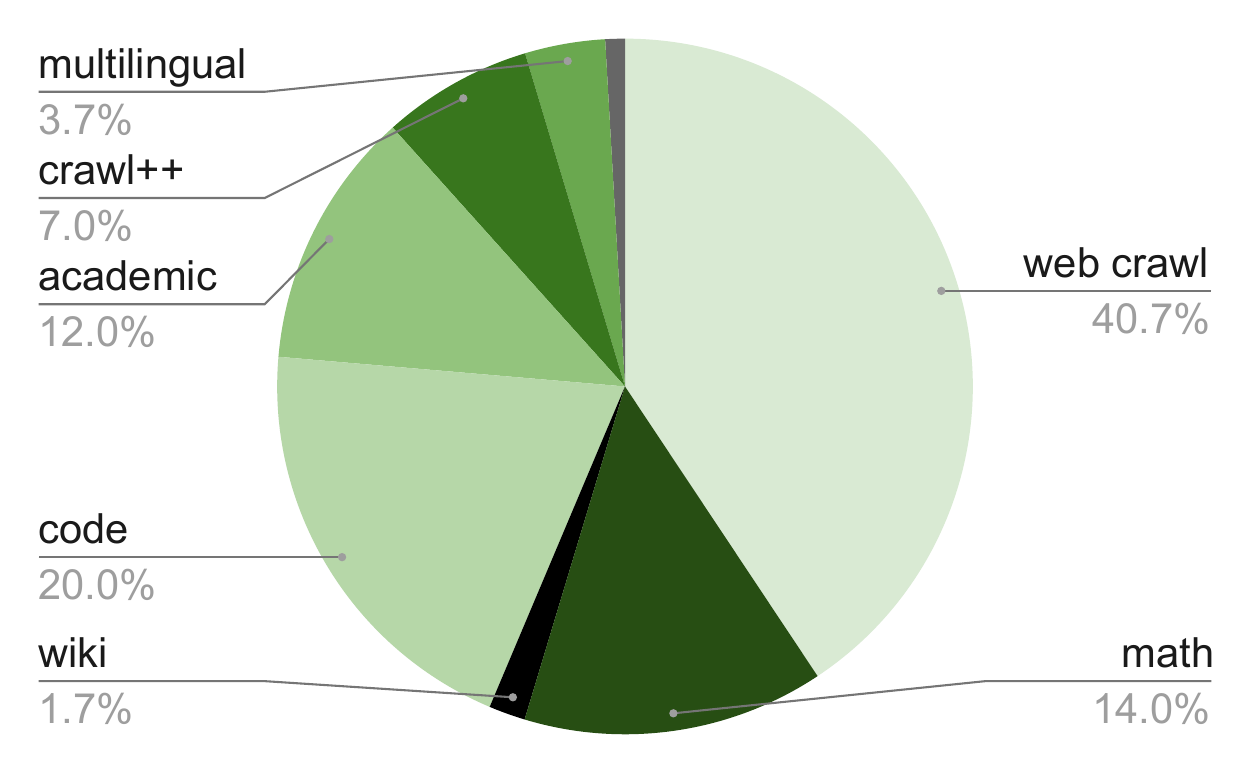}
        \caption{Data mixture of Phase 2.}
        \label{fig:phase2-blend}
    \end{subfigure}
    \vskip\baselineskip
    \begin{subfigure}{0.45\textwidth}
        \centering
        \includegraphics[width=\linewidth]{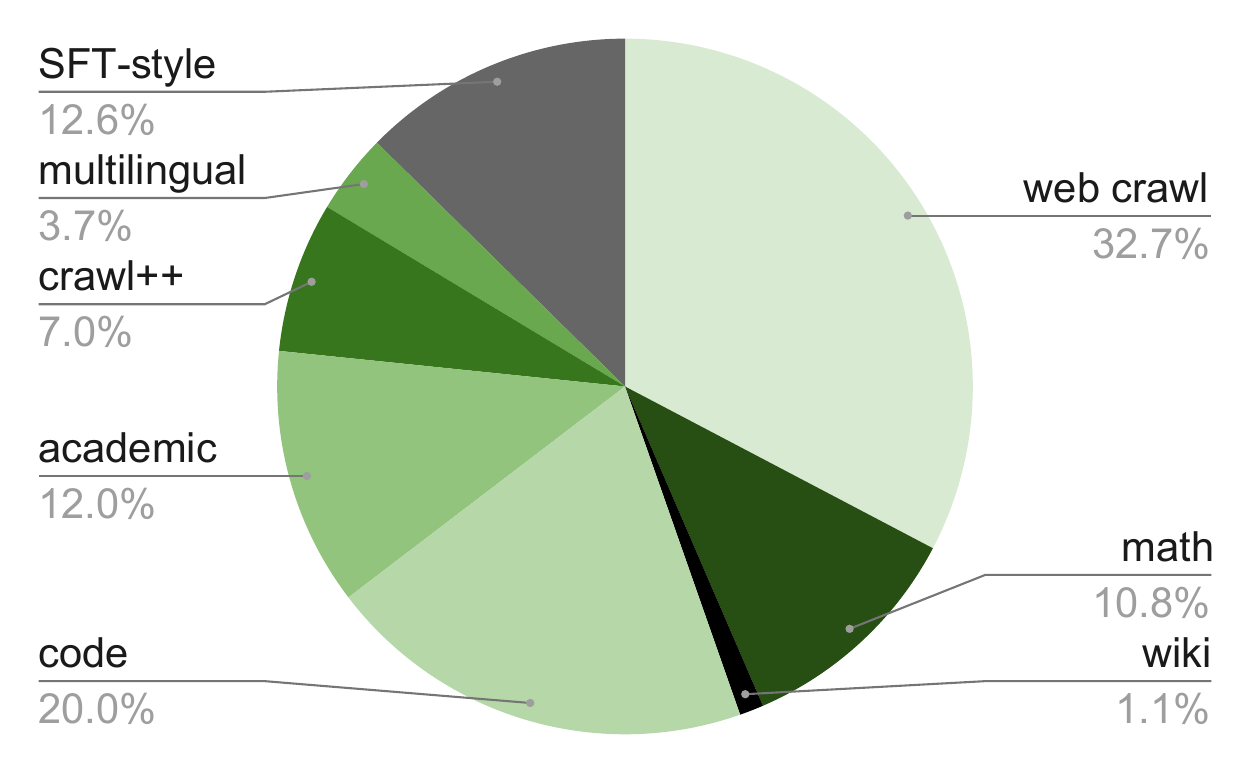}
        \caption{Data mixture of Phase 3.}
        \label{fig:phase3-blend}
    \end{subfigure}
    \hfill
    \begin{subfigure}{0.45\textwidth}
        \centering
        \includegraphics[width=\linewidth]{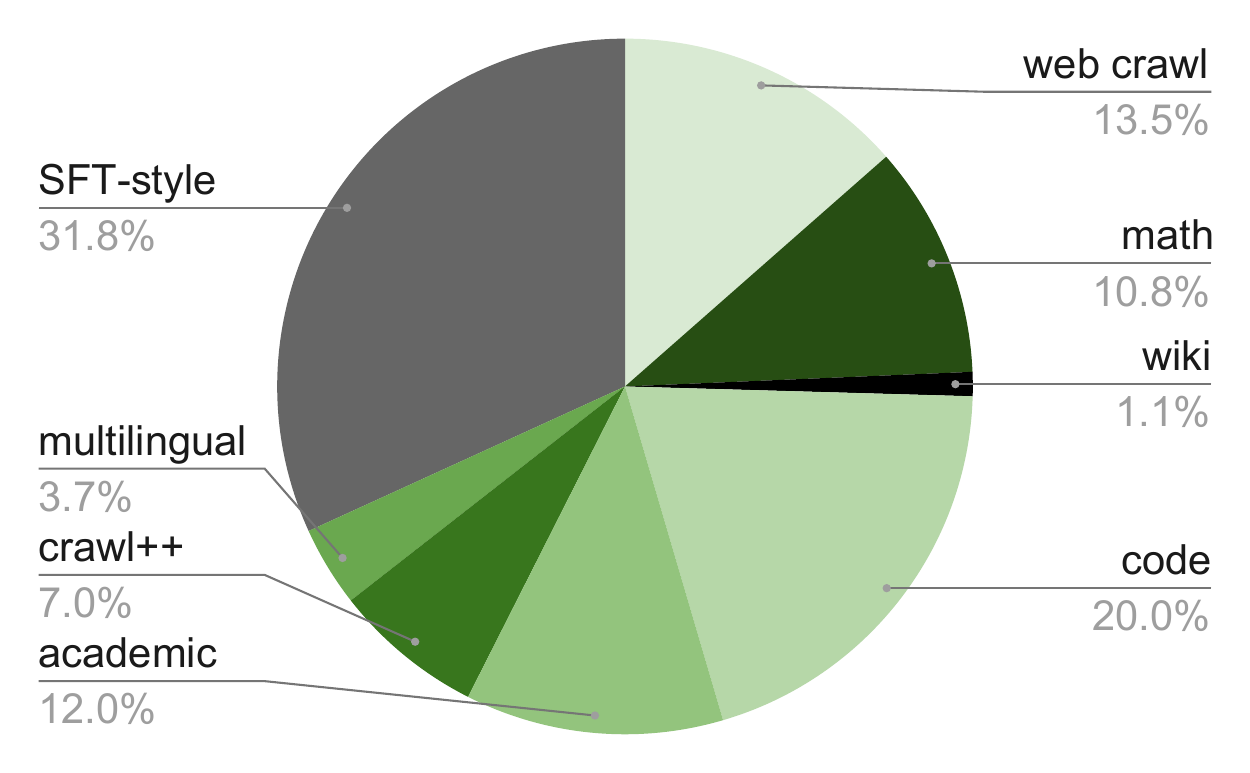}
        \caption{Data mixture of Phase 4.}
        \label{fig:phase4-blend}
    \end{subfigure}
    \caption{Data mixtures for each phase of Nemotron-H pre-training.}
    \label{fig:phase-blends}
\end{figure}

\subsubsection{Data Mixture and Ordering}
\label{section:blend}

Our data mixture consists of seven high-level data categories: web crawl, math, wikipedia, code, academic, crawl++, multilingual, and synthetic SFT-style data.
Crawl++ consists of web-crawl derivatives like OpenWebText, BigScience and Reddit.
Our multilingual data has nine languages: German, Spanish, French, Italian, Portuguese, Chinese, Japanese, Korean, and Russian.
We design the data mixtures in such a way that all data sources of a given quality are weighed similarly, and data sources of higher quality are weighed higher than data sources of lower quality.
We provide more details on estimating the quality of datasets in~\cite{feng2024maximizedataspotentialenhancing}.

We use a phased data-blending approach~\citep{feng2024maximizedataspotentialenhancing} to pre-train both Nemotron-H base models.
In the first phase, we use a data mixture that promotes diversity in data; in the second and third phases, we primarily use high-quality datasets (e.g., Wikipedia). We switch to the second phase at the 60\% point of training, and to the third phase at the 80\% point of training. The fourth phase is performed for the last 380 billion training tokens.
The data mixtures used in each phase are shown in Figure~\ref{fig:phase-blends}.
Our initial experiments on a 8-billion-parameter model trained on 1 trillion tokens show that a phased approach for pre-training outperforms random data ordering by 3.4\%.
\subsection{FP8 Recipe}

Nemotron-H-56B-Base was pre-trained using layer-wise mixed precision: all linear layers in the model (including both linear layers in FFN blocks and the QKV / output projection in the attention block) were computed in FP8 precision, except the first 4 and last 4 layers, which were kept in BF16. We quantized both the forward and backward passes of the linear layers. As in~\cite{micikevicius2022fp8formatsdeeplearning}, we use a hybrid FP8 approach which uses E4M3 for weight and activation tensors, and E5M2 for the gradient tensors. 

We used FP8 per-tensor dynamic quantization, which has several steps: we first compute a quantization scale (with a check for division by zero); then, we multiply the input tensor with the quantization scale and cast it to the desired FP8 format. The quantization scale is computed as a ratio of the maximum representable value for that FP8 data format divided by the maximum absolute value of the input tensor.

\begin{figure}[t]
\centering
\includegraphics[width=0.93\linewidth]{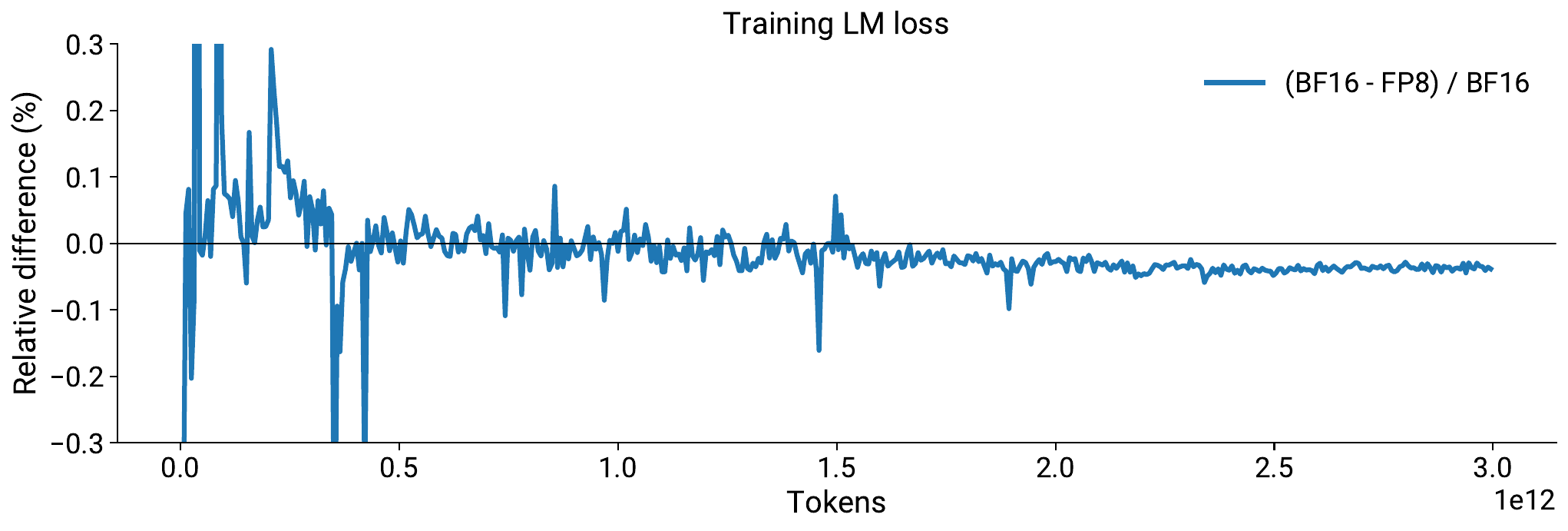}
\caption{The relative training loss gap between a pair of 8B-parameter Transformer models trained for 3 trillion tokens in BF16 and FP8. The gap is initially large, then shrinks, and then grows again in the last stage of training.}
\label{fig:fp8_bf16_diff}
\end{figure}

We observed that the log-likelihood loss curves for FP8 pre-training experiments were almost always higher than for their BF16 counterparts, but by a surprisingly small amount, with typical relative gaps less than 0.1\% on both training and validation. This loss gap tended to decrease as training progressed but then widened in the last quarter of training. We conjecture that this is because extremely small gradient updates, which dominate the latter parts of training, get flushed to zero despite operating in E5M2. Figure~\ref{fig:fp8_bf16_diff} shows a typical relative loss gap between a BF16 and FP8 experiment during pre-training.

Despite the marginally worse loss curves, we observed that downstream evaluations for models trained with FP8 were as good or better than BF16 experiments. Log likelihood, which has traditionally served as a proxy for downstream task, was not a reliable predictor in our experiments; in fact, we often observed models with better loss curves to perform worse on downstream evaluation tasks. 

\begin{figure}[t]
\centering
\includegraphics[width=0.93\linewidth]{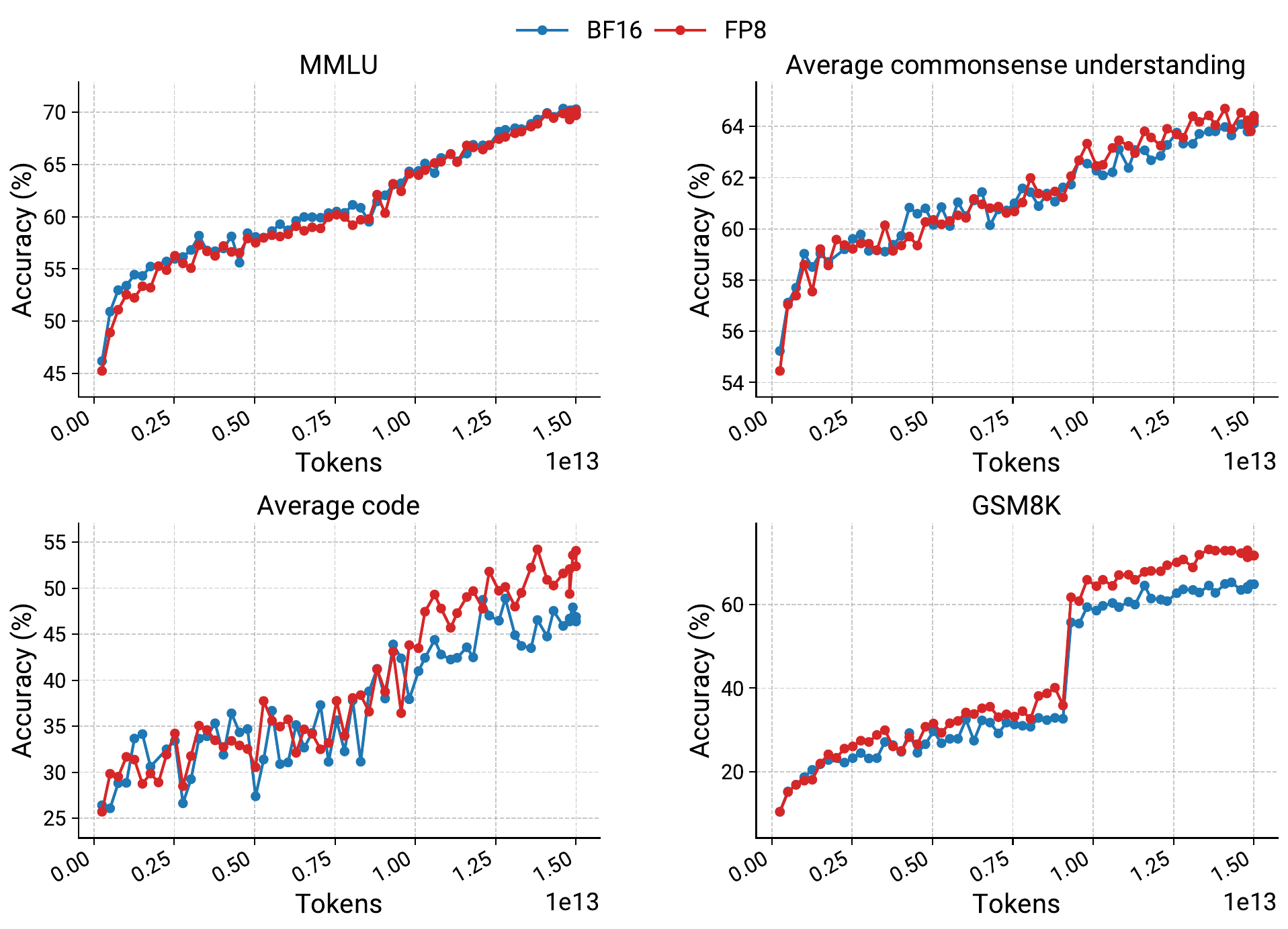}
\caption{Downstream task accuracies for a pair of 8B-parameter Transformer models trained on 15 trillion tokens with BF16 and mixed precision FP8. FP8 accuracies are consistently equal to or better than BF16 ones. The jump in accuracies at 9 trillion tokens is a result of switching data mixtures from Phase 1 to Phase 2.}
\label{fig:fp8_bf16_eval}
\end{figure}

For coding and math tasks, experiments using per-tensor current scaling often had results that were substantially better than BF16 experiments. Figure~\ref{fig:fp8_bf16_eval} shows the accuracy gap between two 8B-parameter Transformers trained for 15 trillion tokens, with one trained in FP8 and the other in BF16. Because of the equal or better downstream results on the same token horizon, we never had to overtrain FP8 models relative to BF16 ones. We ran evaluations in both BF16 and FP8 precision for models trained in FP8; they were typically the same, with BF16 evaluations usually slightly better. However, for the 56B model, the FP8 evaluations eventually outperformed the BF16 ones.

We found it very important to do verification on a minimum of 8B parameters when constructing our FP8 recipe, as results with smaller models did not generalize. All experiments used token horizons of at least 1 trillion tokens.

\subsection{Pre-Training}

We trained Nemotron-H-8B-Base on a token horizon of 15 trillion tokens and Nemotron-H-56B-Base on a token horizon of 20 trillion tokens. We used a sequence length of 8192 and global batch size of 768 (6291456 tokens per batch). We do not use any batch size ramp-up. For Nemotron-H-8B-Base, we used a peak learning rate of $8\cdot10^{-4}$ and warmup over 8.3 billion tokens; for Nemotron-H-56B-Base, we used a peak learning rate of $4\cdot10^{-4}$ and warmup over 25 billion tokens. In both cases, we used cosine learning rate decay with a minimum value equal to 1\% of the peak value, weight decay of 0.1, and 0.9 and 0.95 for Adam $\beta_1$ and $\beta_2$, respectively

We pre-train Nemotron-H using Megatron-LM\footnote{\url{https://github.com/nvidia/megatron-lm}.}; we rely on Transformer Engine\footnote{\url{https://github.com/nvidia/transformerEngine}.} for FP8 support. We use 8-way tensor model parallelism~\citep{shoeybi2020megatronlmtrainingmultibillionparameter} with sequence parallelism~\citep{korthikanti2022reducingactivationrecomputationlarge} for additional memory savings, and 768-way data parallelism with optimizer state distributed over the data-parallel replicas~\citep{rajbhandari2020zeromemoryoptimizationstraining}.

\subsubsection{Resiliency}

We also focus heavily on minimizing downtime when training models at scale: over long training jobs on a large number of GPUs, job failures and slowdowns are inevitable, leading to lost training productivity. Failures can occur due to GPU-related errors (e.g., uncorrectable ECC errors or GPUs falling off the bus), errors in shared infrastructure like the network fabric or Lustre, or application-induced failures (e.g., numerical instability in certain operations or ``bad'' data). We can also experience job slowdowns due to a variety of reasons (e.g., single-GPU throttling or slower communication collectives). 

To minimize the impact of the above issues, we focus on a) accurately attributing failures to the correct infrastructure and application components to prevent repeated disruptions, and b) reducing the recovery overhead of each failure. To do so, we leverage the DGX Cloud Resilience service\footnote{\url{https://developer.nvidia.com/blog/ensuring-reliable-model-training-on-nvidia-dgx-cloud/}.} to ensure failures are correctly attributed and the appropriate hardware component (e.g., particular HGX node with failing GPU) is not included in the next instance of the training job, preventing repeated failures. With DGX Cloud Resilience enabled, we obtained a 3.3$\times$ improvement in mean time between failures (MTBF) in our cluster.

We also proactively save checkpoints frequently in order to minimize the amount of work lost when a failure occurs (weight updates from the last checkpoint save to the failure are lost). We use NVRx\footnote{\url{https://github.com/NVIDIA/nvidia-resiliency-ext}.} to facilitate checkpointing with low overhead by asynchronously saving checkpoints without blocking training. We also optimized job restart times (i.e., the set of operations needed by Megatron-LM before training iterations can start like distributed runtime initialization and checkpoint loading).

Altogether, the above components enable us to train Nemotron-H-56B-Base with high efficiency on 6144 NVIDIA H100 GPUs.

\subsection{Base Model Evaluations}
\label{subsec:base_model_evals}

We run evaluations of all models ourselves unless otherwise stated. Our evaluation setup is built on top of \texttt{lm-evaluation-harness}\footnote{\url{https://github.com/EleutherAI/lm-evaluation-harness}.} for fair comparisons, with the following changes:
\begin{enumerate}
    \item For mathematical reasoning, we evaluate on the GSM8K and MATH~\citep{cobbe2021trainingverifierssolvemath, hendrycks2021measuringmathematicalproblemsolving} benchmarks. We also highlight the competition-level slice of the MATH benchmark as ``MATH Level 5''. After carefully looking through disagreeing cases, we use a combination of \texttt{Math-Verify}\footnote{\url{https://github.com/huggingface/math-verify}.} and the math grading utilities of \texttt{NeMo-Skills}\footnote{\url{https://github.com/NVIDIA/NeMo-Skills/blob/main/nemo_skills/code_execution/math_grader.py}.} to grade solutions.
    \item We also include a MATH-R1 version of the MATH task, where we provide few-shot examples generated by DeepSeek-R1~\citep{deepseekai2025deepseekr1incentivizingreasoningcapability}. Both Nemotron-H and the Qwen 2.5 models show significant improvement on the benchmark by simply changing the prompt.
    \item For code tasks (HumanEval~\citep{chen2021evaluatinglargelanguagemodels}, MBPP~\citep{austin2021programsynthesislargelanguage} and their EvalPlus variants), we always sanitize the generations using EvalPlus~\citep{Liu_Is_Your_Code_2023}.
    \item General reasoning benchmarks (OpenBookQA~\citep{mihaylov2018suitarmorconductelectricity}, PIQA~\citep{bisk2019piqareasoningphysicalcommonsense}, Hellaswag~\citep{zellers2019hellaswagmachinereallyfinish}, Winogrande~\cite{sakaguchi2019winograndeadversarialwinogradschema}) are unchanged except for ARC-Challenge~\citep{Clark2018ThinkYH}, where we present all options at the same time, similar to MMLU~\citep{hendrycks2021measuringmassivemultitasklanguage}.
\end{enumerate}

\begin{table}[t]\scriptsize
    \centering
    \renewcommand{\arraystretch}{1.2}
    \setlength{\tabcolsep}{6pt}
    \begin{tabularx}{\textwidth}{l *{6}{>{\centering\arraybackslash}X}}
        \toprule
        \textbf{Task} &
        \textbf{\begin{tabular}[c]{@{}c@{}}{N-H}\\ 56B-Base\end{tabular}} &
        \textbf{\begin{tabular}[c]{@{}c@{}}{N-H}\\ 47B-Base\end{tabular}} &
        \textbf{\begin{tabular}[c]{@{}c@{}}Qwen-2.5 \\ 72B-Base\end{tabular}} &
        \textbf{\begin{tabular}[c]{@{}c@{}}Llama-3.1\\ 70B-Base\end{tabular}} &
        \textbf{\begin{tabular}[c]{@{}c@{}}DS-V3 \\ 671B-Base\end{tabular}} &
        \textbf{\begin{tabular}[c]{@{}c@{}}Llama-3.1\\ 405B-Base\end{tabular}} \\ \toprule
        \multicolumn{7}{l}{\textbf{General}} \\
        MMLU-Pro (5-shot COT) & 60.5 & \textbf{61.8} & 58.8 & 51.3 & \textit{64.4} & \textit{52.8} \\
        MMLU (5-shot) & 84.2 & 83.6 & \textbf{86.1} & 78.8 & \textit{87.1} & \textit{84.4} \\
        \midrule
        \multicolumn{7}{l}{\textbf{Math}} \\
        GSM8k (8-shot COT) & \textbf{93.7} & 93.3 & 90.9 & 83.9 & \textit{89.3} & \textit{83.5} \\
        MATH (4-shot COT) & 59.4 & 57.4 & \textbf{64.6} & 42.9 & \textit{61.6} & \textit{49.0} \\
        MATH Level 5 (4-shot COT) & 35.2 & 34.1 & \textbf{41.2} & 21.1 & & \\
        MMLU STEM (5-shot) & 80.6 & 79.8 & \textbf{84.9} & 70.5 & & \\
        \midrule
        \multicolumn{7}{l}{\textbf{Code}} \\
        HumanEval (0-shot greedy pass@1) & 60.4 & \textbf{61.0} & 56.7 & 57.3 & \textit{65.2} & \textit{54.9} \\
        HumanEval+ (0-shot greedy pass@1) & 54.3 & \textbf{56.1} & 50.0 & 52.4 & & \\
        MBPP sanitized (3-shot greedy pass@1) & 77.8 & 75.9 & \textbf{78.2} & 70.0 & \textit{75.4} & \textit{68.4} \\
        MBPP+ (0-shot greedy pass@1) & 67.2 & 65.6 & \textbf{71.7} & 66.9 & & \\
        \midrule
        \multicolumn{7}{l}{\textbf{Commonsense understanding}} \\
        Arc-Challenge (25-shot) & 95.0 & 94.6 & \textbf{95.8} & 93.0 & \textit{95.3} & \textit{95.3} \\
        Hellaswag (10-shot) & \textbf{89.0} & 87.9 & 87.6 & 88.1 & \textit{88.9} & \textit{89.2} \\
        Winogrande (5-shot) & 84.5 & 83.9 & 84.4 & \textbf{85.6} & \textit{84.9} & \textit{85.2} \\
        PIQA (0-shot) & \textbf{85.0} & 83.9 & 83.6 & 84.1 & \textit{84.7} & \textit{85.9} \\
        OpenbookQA (0-shot) & 48.6 & \textbf{48.8} & 46.2 & 47.6 & & \\
        \midrule
        \multicolumn{7}{l}{\textbf{Long-thought reasoning}} \\
        MATH (R1-style 4-shot COT) & \textbf{87.8} & & 73.9 & 38.7 & & \\
        MATH Level 5 (R1-style 4-shot COT) & \textbf{74.8} & & 53.2 & 16.7 & & \\
        \bottomrule
    \end{tabularx}
    \caption{Accuracy of Nemotron-H-56B-Base (and its distilled version Nemotron-H-47B-Base, discussed in \S\ref{sec:distillation_full}) versus existing SoTA models. N-H is short for Nemotron-H, DS is short for DeepSeek. Numbers in italics are from the DeepSeek-V3 report~\citep{deepseekai2025deepseekv3technicalreport}, all other numbers are run by us. We bold the highest accuracy in each row, excluding italicized numbers.}
    \label{tab:accuracy_comparison_56b}
\end{table}

Accuracy results for Nemotron-H-56B-Base and Nemotron-H-8B-Base on common benchmark tasks are shown in Table~\ref{tab:accuracy_comparison_56b} and Table~\ref{tab:accuracy_comparison_8b}. Nemotron-H-56B-Base and Nemotron-H-47B-Base are both evaluated in FP8, and all other models are evaluated in BF16. Across both model sizes, Nemotron-H base models reach comparable or better accuracy relative to similarly-sized state-of-the-art Transformer models.

We compare Nemotron-H-56B-Base directly with Qwen-2.5-72B-Base~\citep{qwen2025qwen25technicalreport} and Llama-3.1-70B-Base~\citep{grattafiori2024llama3herdmodels}. Out of the 17 task we evaluate, Nemotron-H-56B-Base achieves the highest accuracy of the three models on 9 tasks, Qwen-2.5-72B-Base achieves the highest accuracy on 7 tasks, and Llama-3.1-70B-Base achieves the highest accuracy on 1 task; Nemotron-H-56B-Base outperforms Llama-3.1-70B-Base on all but that single task (Winogrande). We further compare Nemotron-H-56B-Base with two much larger state-of-the-art models, DeepSeek-V3-671B-Base~\citep{deepseekai2025deepseekv3technicalreport} and Llama-3.1-405B-Base~\citep{grattafiori2024llama3herdmodels}; we use publicly-reported numbers for these two models on the subset of common tasks where numbers are reported. Surprisingly, Nemotron-H-56B-Base remains competitive with these models, outperforming DeepSeek-V3-671B-Base and Llama-3.1-405B-Base on 4 and 5 out of the 10 overlapping tasks respectively.

\begin{table}[t]\scriptsize
    \renewcommand{\arraystretch}{1.2}
    \setlength{\tabcolsep}{6pt}
    \centering
    \begin{tabularx}{\textwidth}{l *{6}{>{\centering\arraybackslash}X}}
        \toprule
        \textbf{Task} &
        \textbf{\begin{tabular}[c]{@{}c@{}}{Nemotron-H}\\ 8B-Base\end{tabular}} &
        \textbf{\begin{tabular}[c]{@{}c@{}}{Nemotron-T}\\ 8B-Base\end{tabular}} &
        \textbf{\begin{tabular}[c]{@{}c@{}}Qwen-2.5 \\ 7B-Base\end{tabular}} &
        \textbf{\begin{tabular}[c]{@{}c@{}}Llama-3.1\\ 8B-Base\end{tabular}} &
        \textbf{\begin{tabular}[c]{@{}c@{}}Gemma-3 \\ 12B-Base\end{tabular}} \\ \toprule
        \multicolumn{6}{l}{\textbf{General}} \\
        MMLU-Pro (5-shot COT) & 44.0 & 41.4 & \textbf{48.3} & 35.9 & 45.3 \\
        MMLU (5-shot) & 72.8 & 73.2 & \textbf{74.2} & 65.3 & \textit{74.5} \\
        \midrule
        \multicolumn{6}{l}{\textbf{Math}} \\
        GSM8k (8-shot COT) & 87.1 & \textbf{89.0} & 83.3 & 55.5 & 74.1 \\
        MATH (4-shot COT) & 46.5 & 46.7 & \textbf{49.8} & 19.5 & 42.1 \\
        MATH Level 5 (4-shot COT) & 22.9 & \textbf{25.8} & 24.6 & 5.6 & 17.5 \\
        MMLU STEM (5-shot) & 65.4 & 65.6 & \textbf{71.2} & 56.3 &  \\
        \midrule
        \multicolumn{6}{l}{\textbf{Code}} \\
        HumanEval (0-shot greedy pass@1) & 58.5 & \textbf{59.8} & 56.7 & 37.8 & 46.3 \\
        HumanEval+ (0-shot greedy pass@1) & \textbf{55.5} & 53.0 & 48.8 & 31.7 & 34.1 \\
        MBPP sanitized (3-shot greedy pass@1) & 65.4 & 63.4 & \textbf{69.3} & 57.2 & 64.6 \\
        MBPP+ (0-shot greedy pass@1) & 59.5 & 61.4 & \textbf{65.3} & 51.6 & 59.0 \\
        \midrule
        \multicolumn{6}{l}{\textbf{Commonsense understanding}} \\
        Arc-Challenge (25-shot) & 88.7 & 88.3 & \textbf{89.2} & 80.4 &  \\
        Hellaswag (10-shot) & \textbf{83.2} & 82.5 & 80.3 & 82.3 & \textit{84.2} \\
        Winogrande (5-shot) & \textbf{80.5} & 78.8 & 76.1 & 78.1 & \textit{74.3} \\
        PIQA (0-shot) & \textbf{82.2} & 82.0 & 80.1 & 81.0 & \textit{81.8} \\
        OpenbookQA (0-shot) & \textbf{47.2} & 44.8 & 47.0 & 45.4 &  \\
        \bottomrule
    \end{tabularx}
    \caption{Accuracy of Nemotron-H-8B-Base versus existing SoTA models and a Transformer (Nemotron-T-8B-Base) trained on exactly the same data. Likelihood-based evaluations of the Gemma 3 model are based on the Gemma 3 report~\citep{gemmateam2025gemma3technicalreport}, all other numbers are run by us. We bold the highest accuracy in each row, excluding italicized numbers.}
    \label{tab:accuracy_comparison_8b}
\end{table}

We observe similar accuracy results for Nemotron-H-8B-Base compared to Qwen-2.5-7B-Base and Llama-3.1-8B-Base (Table~\ref{tab:accuracy_comparison_8b}). Out of the 15 tasks used for evaluation, Nemotron-H-8B-Base, Qwen-2.5-7B-Base, and Llama-3.1-8B-Base achieve the highest accuracy of the three models on 7, 8, and 0 tasks respectively. Nemotron-H-8B-Base is particularly strong on commonsense understanding tasks, achieving the highest accuracy on 4 out of 5 tasks in this category. As above, we also include comparisons to a larger model, Gemma-3-12B-Base~\citep{gemmateam2025gemma3technicalreport}; Nemotron-H-8B-Base is competitive with the larger model. 

Table~\ref{tab:accuracy_comparison_8b} additionally includes the largest apples-to-apples comparison of a hybrid Mamba-Transformer and pure Transformer model to date (Nemotron-H-8B-Base versus Nemotron-T-8B-Base). Nemotron-H-8B-Base reaches higher accuracy than the Transformer trained on exactly the same data on 7 out of 15 tasks, and is within one point on an additional 4 tasks. Overall, this experiment shows that hybrid models can reach equal or better accuracy compared to Transformer models when trained at state-of-the-art scales. 

\begin{figure}[t]
    \centering
    \includegraphics[width=0.93\linewidth]{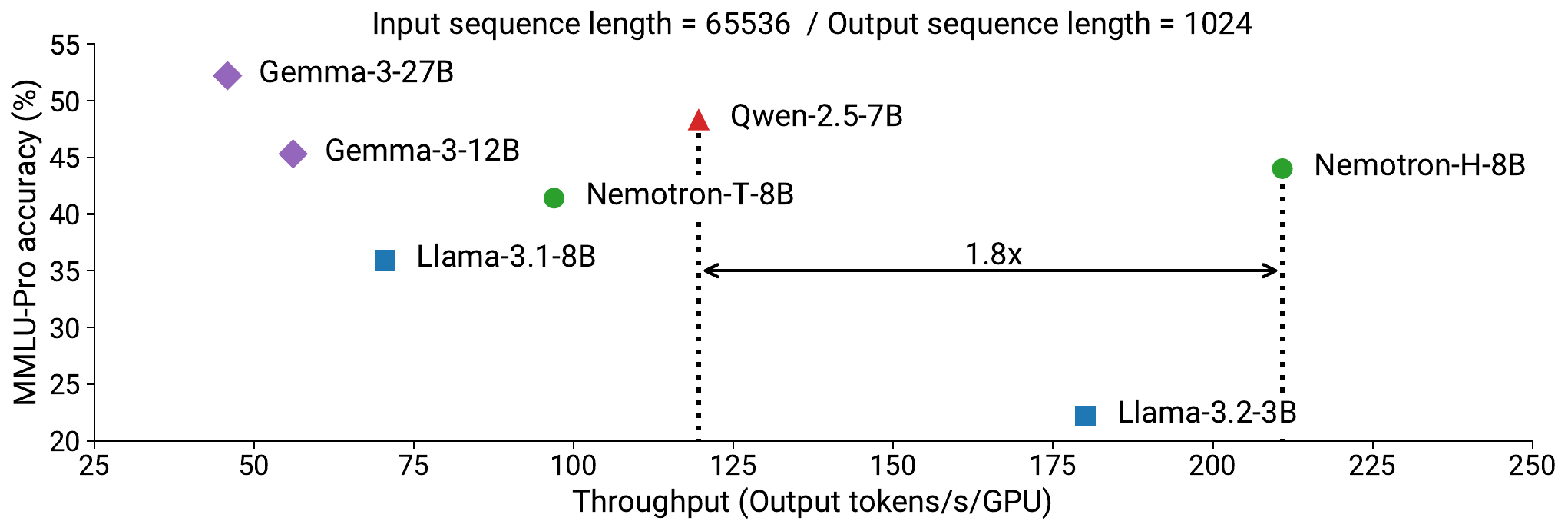}
    \caption{MMLU-Pro accuracy versus inference throughput (normalized by number of GPUs used) for Nemotron-H-8B-Base compared to existing similarly-sized Transformer models.}
    \label{fig:inference_speed_8b}
\end{figure}

\subsection{Inference Speed}
Due to the reduction in self-attention layers, Nemotron-H-8B/56B-Base provide inference-time speedups compared to the alternative Transformer models in Tables~\ref{tab:accuracy_comparison_56b} and~\ref{tab:accuracy_comparison_8b} above. To quantify these speedups, we plot the MMLU-Pro accuracy versus inference throughput for Nemotron-H-56B-Base and similarly-sized Transformer models in Figure~\ref{fig:inference_speed_56b} and for Nemotron-H-8B-Base and relevant baselines in Figure~\ref{fig:inference_speed_8b}. We use an input sequence length of 65536 and ask the models to generate 1024 output tokens. We use an initial Megatron-LM implementation for Nemotron-H inference and vLLM v0.7.3\footnote{\url{https://github.com/vllm-project/vllm/tree/v0.7.3}.} for baselines. In these experiments, we try to maximize per-GPU inference throughput by using as large a batch size as possible, and we run all experiments on NVIDIA H100 GPUs. We report results normalized by the number of GPUs used (i.e., output tokens per second per GPU).

In the setting described above, Nemotron-H-56B-Base can generate 2.4$\times$ more output tokens per second per GPU compared to Qwen-2.5-72B and Llama-3.1-70B (and 2.9$\times$ more after distillation to Nemotron-H-47B-Base, see \S\ref{sec:distillation_full}). Compared to Llama-3.1-405B, Nemotron-H-56B-Base achieves 19.6$\times$ higher throughput. Similar to the larger models, Nemotron-H-8B-Base is also faster to infer than corresponding Transformer models. As shown in Figure ~\ref{fig:inference_speed_8b}, on longer contexts, we measure a 1.8$\times$ and 3$\times$ speedup compared to Qwen-2.5-7B and Llama-3.1-8B. Inference speedups are a byproduct of two factors: a) constant computation in the Mamba-2 layers as opposed to linear computation in self-attention layers, b) lower memory footprint from Mamba-2 layers facilitates using higher batch sizes, increasing efficiency. We expect further optimizing Nemotron-H inference to lead to additional speedups compared to Transformers, which have been heavily optimzed over the last couple of years.

\section{Compression and Distillation}
\label{sec:distillation_full}

Efficient deployment of LLMs often requires tailoring the model architecture to specific constraints imposed by hardware.
In this section, we describe the pruning and distillation process used to compress the 56B model to 47B parameters with the goal of running longer context inference on the NVIDIA GeForce RTX 5090 GPU (storing the weights of a 56B parameter model in FP4 precision on a RTX 5090 GPU with 32GiB of memory will require 29.5GiB, leaving only 2.5GiB for KV cache and activation buffers). We introduce a novel model compression framework called MiniPuzzle that combines the simplicity of Minitron~\citep{sreenivas2024llmpruningdistillationpractice,muralidharan2024compactlanguagemodelspruning} with the versatility of Puzzle~\citep{bercovich2024puzzledistillationbasednasinferenceoptimized}. 
Our approach uses roughly 300$\times$ fewer tokens to obtain a 47B model compared to training from scratch. 
Our compressed 47B model achieves a 1.2$\times$ inference speedup (Figure~\ref{fig:inference_speed_56b}) while achieving accuracy on par with the original 56B model.

\subsection{MiniPuzzle Overview}

\begin{figure*}[thbp]
    \centering
    \includegraphics[width=\linewidth]{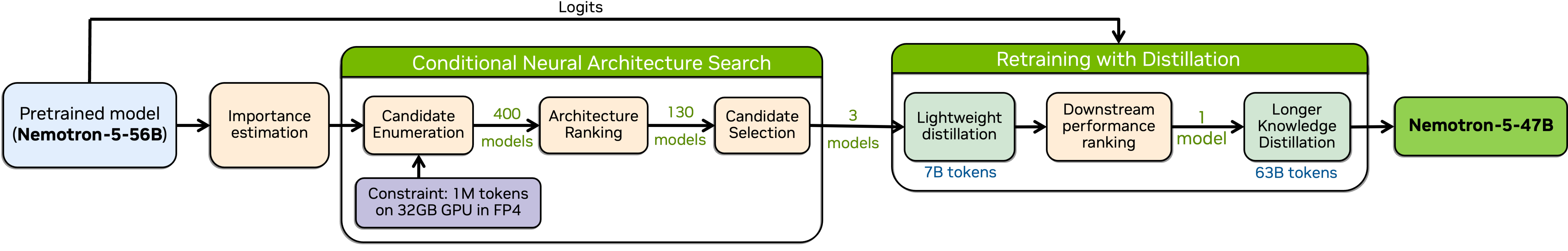}
    \caption{MiniPuzzle's optimization process: a) estimate layer and FFN importance scores, b) search for the best model candidates using the conditional Neural Architecture Search (NAS) framework, considering the 32GiB GPU memory constraint, c) select the best candidate with lightweight distillation and recover the accuracy lost due to pruning with longer distillation.}
    \label{fig:minipuzzle_overview}
\end{figure*}

MiniPuzzle combines lightweight pruning with neural architecture search (NAS) to produce efficient compressed models that meet specific target deployment constraints.
Figure~\ref{fig:minipuzzle_overview} provides a high-level overview of the framework. MiniPuzzle's optimization process consists of three stages: a) importance estimation (\S\ref{sec:importance}), b) conditional NAS (\S\ref{section:search}), and c) knowledge distillation (\S\ref{sec:distillation}).

\subsection{Importance Estimation}
\label{sec:importance}
MiniPuzzle first collects importance or sensitivity scores for each model component (e.g., layers, FFN neurons) to help decide which components to remove; this is the {\em importance estimation} phase.
The scores computed in this phase are used to decide which model components can be pruned.
We note that sensitivity analysis based on gradient information is typically impractical at modern LLM scale~\citep{muralidharan2024compactlanguagemodelspruning};
instead, we rely on a lightweight strategy that uses only on forward passes. In this work, we use a simplified approach that works well in our ablation studies: a) prune layers, and b) prune FFN hidden dimensions (effectively neurons).
We also experimented with pruning Mamba heads and Mamba head dimension; unfortunately, both axes caused severe accuracy degradation (particularly head dimension pruning).

We now describe how we compute the importance of each layer and FFN neuron.

\paragraph{Layer importance.}
We use the scoring method from Puzzle to estimate the importance of each layer in the model. Given an input tensor, layer importance is computed as the Mean Squared Error (MSE) between the intermediate activation tensor just before the LM head of the full model and the same activation tensor for a model with the particular layer removed.
We average these rankings over a small random subset of the training data (128 samples in our case) to obtain a reliable estimate of importance that takes into account sample variability.

\begin{figure}[!ht]
\centering
\includegraphics[width=0.9\linewidth]{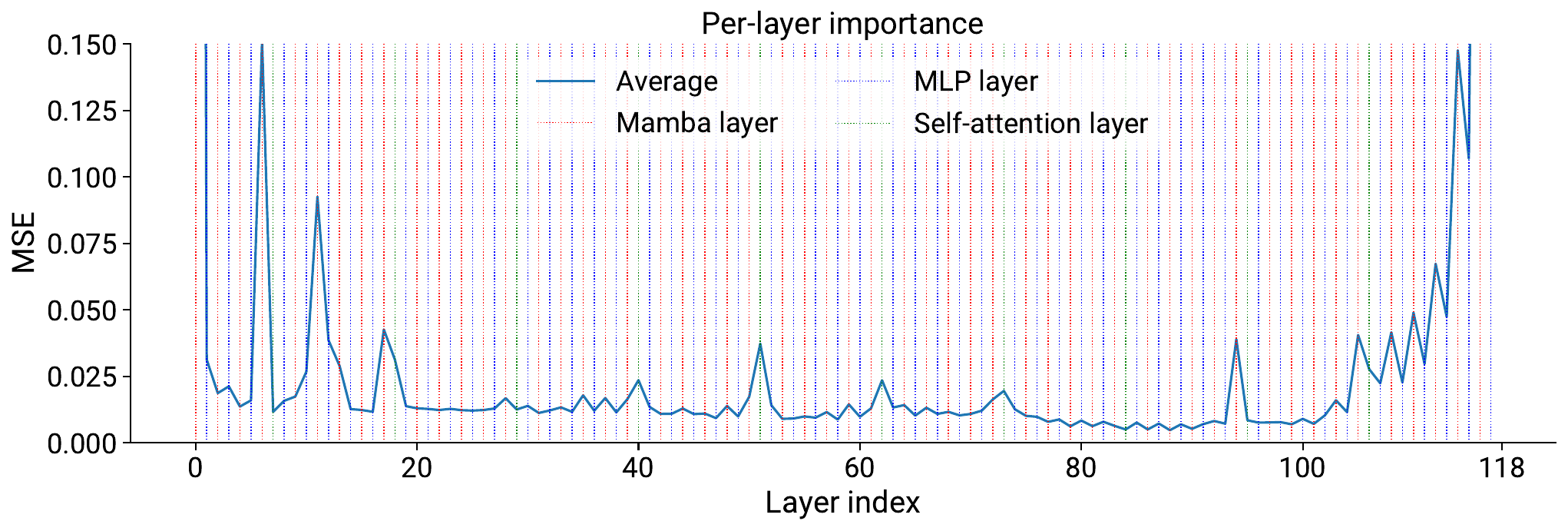}
\caption{Layer importance measured as the MSE between the full model and a model with that layer removed, averaged over a small training subset. Vertical dotted lines indicate layer types: self-attention (green), FFN (blue), and Mamba-2 (red).
}
\label{fig:per-layer-importance-MSE}
\end{figure}

Figure~\ref{fig:per-layer-importance-MSE} plots average importance scores of each layer in Nemotron-H-56B-Base. The green, blue and red dotted lines correspond to self-attention, FFN and Mamba layers.
We notice from the figure that the 56B model follows a typical pattern observed in Minitron:
the most important layers are concentrated at the beginning and end of the model. Additionally, MSE importance reveals that even though the 56B model only has 10 self-attention layers, some self-attention layers are ranked among the least important, particularly the 84th (7th self-attention layer). However, layers 40, 51, 62 and 73 seem to be more important compared to the other layers in their immediate vicinity. 

\paragraph{FFN importance.}\label{mlp_pruning} FFN layers internally are composed of two linear operators with a non-linear activation in between: 
$$
\operatorname{FFN}(\mathbf{X}) = \delta\bigg(\mathbf{X} \cdot \boldsymbol{W}^{T}_{1}\bigg) \cdot \boldsymbol{W}_{2}.
$$
Here, $\mathbf{X}$ denotes the input, and $\boldsymbol{W}_{1}$ and $\boldsymbol{W}_{2}$ are the two associated weight matrices in the FFN layer. 
$\boldsymbol{W}_{1}, \boldsymbol{W}_{2} \in \mathbb{R}^{d_{ffn}\times d_{model}}$, where $d_{model}$ and $d_{ffn}$ are the model hidden dimension and FFN hidden dimension respectively (Table~\ref{tab:nemotron-h-arch}).
$\delta(\cdot)$ refers to the non-linear activation function (squared ReLU in this work).

Following the same procedure as Minitron, we compute the importance of each neuron in the first linear operator of each FFN layer by examining the set of outputs it produces. We use a small calibration dataset of 1024 samples for this purpose. Formally, we compute each neuron's importance score by aggregating its outputs given an input batch $X$:
$$F_{\text{neuron}}^{(i)} = \sum_{\mathbf{B,S}} \delta\bigg(\mathbf{X} \big(\boldsymbol{W}_{1}^{i}\big)^T\bigg).$$
Here, $\boldsymbol{W}_{1}^{i}$ refers to the $i^\text{th}$ row of the weight matrix $\boldsymbol{W_{1}}$. 
$\sum_{\mathbf{B, S}}$ refers to aggregation along the batch and sequence dimensions. 
We use the \texttt{mean} and \texttt{l2-norm} aggregation functions along the batch and sequence dimensions, following the observations in the Minitron paper~\citep{muralidharan2024compactlanguagemodelspruning}.
For a sequence of scores $\mathbf{S}$, \texttt{mean} aggregation is defined as $\frac{1}{n}\sum_{i=1}^{n}|\mathbf{S}_i|$, and \texttt{l2-norm} is $\sqrt{\sum_{i=1}^{n}\mathbf{S}_i^2}$.

\subsection{Conditional NAS}
\label{section:search}

MiniPuzzle's conditional NAS utilizes the importance scores of every layer and FFN neuron (computed as described in \S\ref{sec:importance}) to identify architectures that meet memory constraints while preserving the most critical components. This process allows us to efficiently explore the vast search space of possible layer configurations, and consists of three steps:

\begin{figure}[!ht]
    \centering
    \includegraphics[width=0.93\linewidth]{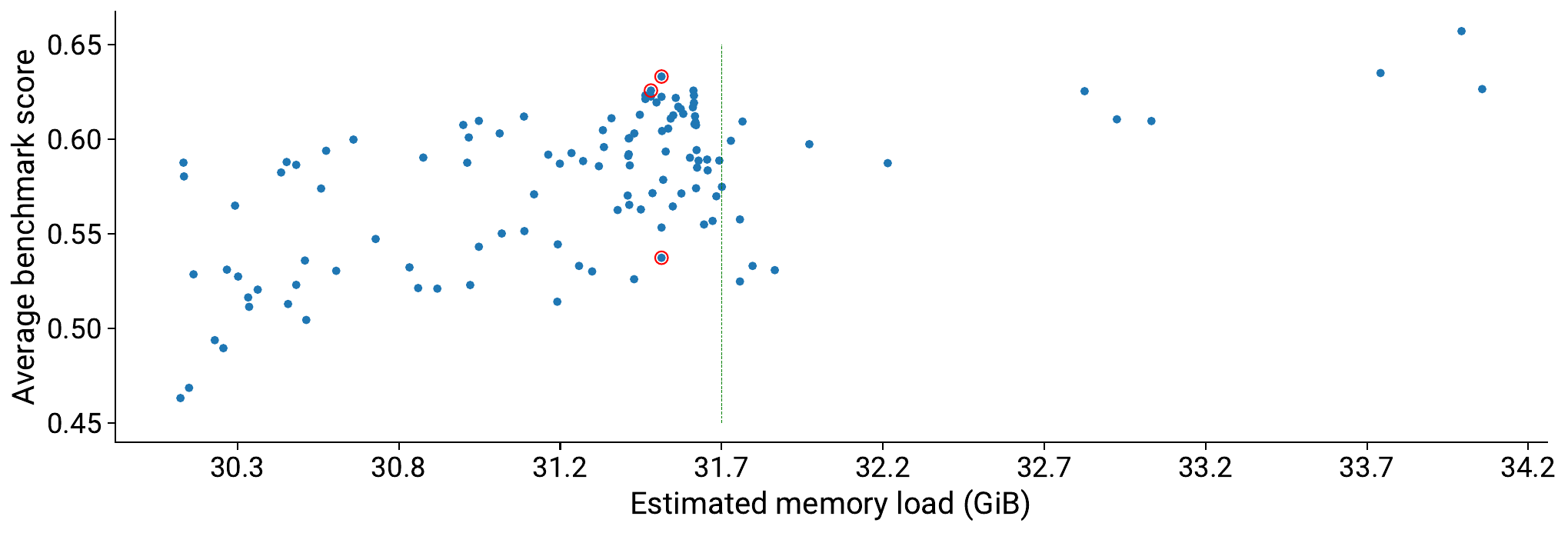}
    \caption{Average benchmark score for pruned candidates with different memory budgets. Each dot represents a different candidate, i.e., model with different set of layers pruned with respect to how much memory it consumes. The green dotted line marks the upper memory limit we consider and the dots marked with red circles are the candidates selected for further distillation.}
    \label{fig:candidates}
\end{figure}

\begin{figure}[!ht]
    \centering
    \includegraphics[width=0.93\linewidth]{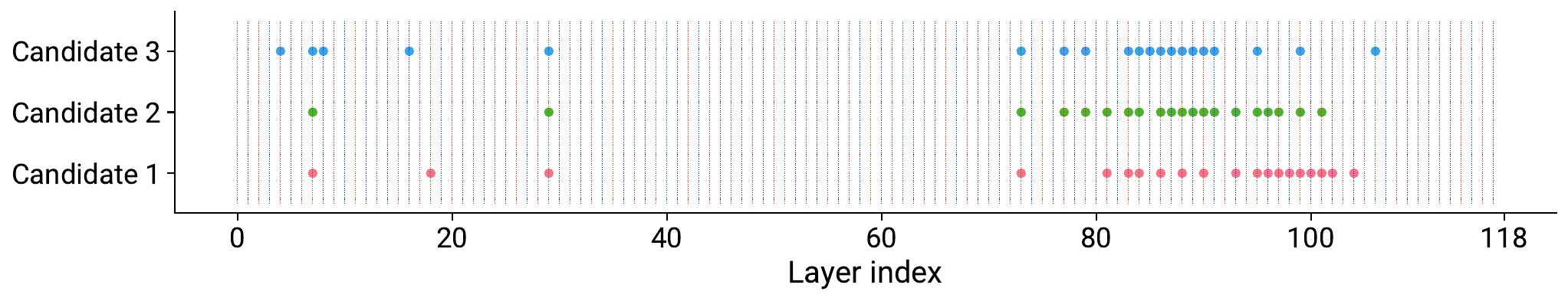}
    \caption{Pruned layers for the top three candidates. Each row represents a candidate, with dots indicating dropped layers. Vertical dotted lines denote layer types: green for self-attention, blue for FFN, and red for Mamba.}
    \label{fig:candidates-selected}
\end{figure}

\begin{enumerate}
    \item \textbf{Pruned candidate enumeration.} We iterate over a grid of possible layer counts for each layer type (i.e., number of self-attention, FFN, and Mamba-2 layers), and FFN hidden dimension (24576, 25600, \ldots, 32768). For each target layer count and FFN hidden dimension size, we select the top layers and neurons based on their respective importance scores (see \S\ref{sec:importance} for details on importance estimation) and realize the corresponding architecture; i.e., we drop the irrelevant layers and prune all FFN layers to the same target width. Only architectures that meet the target memory constraint of less than 31.7 GiB (computed as the total memory required for FP4 inference on a context of size 1 million tokens) are retained, resulting in around 400 candidate architectures.
    \item \textbf{Architecture ranking.} We score the candidate architectures using a lightweight metric that compares them to the parent model (Nemotron-H-56B-Base) on a small validation corpus with 1 million tokens. Specifically, we use two different scores to estimate the quality of each candidate model: a) next-token accuracy which computes how often the child model correctly predicts the next ground-truth token, and b) next-token parent agreement which computes how often the child model agrees with the parent model about the greedy prediction of the next token. We rank all candidate architectures using both scoring metrics, and retain the top 130 performers.
    \item \textbf{Candidate selection.} We now refine the set of ranked architectures from 130 to a manageable 3 by benchmarking them on a subset of evaluation tasks. The first column of Table~\ref{tab:zero_shot_scores} lists the tasks we use to evaluate the final 3 candidates. We use the average scores across all these tasks for evaluation. Figure~\ref{fig:candidates} plots the average benchmark scores for all 130 candidates, and Figure~\ref{fig:candidates-selected} illustrates the specific layers dropped by all three candidates. Candidates 1 and 2 achieve the highest benchmark scores and drop similar layers. Candidate 3, which drops a slightly different set of layers, performs significantly worse after pruning. However, we include it here to study the extent to which distillation can compensate for the loss.
\end{enumerate}
  \noindent

\subsection{Retraining with Distillation}
\label{sec:distillation}
To recover the accuracy lost due to pruning, the model undergoes continued training. Recent work has demonstrated that distilling knowledge from the original model to the pruned model outperforms conventional fine-tuning (as shown in Minitron and Puzzle);
we thus adopt logit-based distillation for continued training, employing forward KL divergence loss exclusively during the accuracy recovery phase (see \S3 of the Minitron paper for more details on the distillation loss formulation).

Following the initial candidate selection process described in \S\ref{section:search}, we perform short distillation of the top 3 candidates using $\sim$7B tokens, with the 56B model as the teacher. We observe from our experiments that this step is crucial for picking the best possible final candidate. Table~\ref{tab:zero_shot_scores} details the benchmark scores for the 3 candidates before and after short distillation. The most accurate candidate (averaged across all benchmarks) of the three is chosen for an extended distillation run using 63 billion tokens to produce the final 47B model; in our case, Candidate 2 performs best, as shown in Table~\ref{tab:zero_shot_scores}.
We observe that the choice of the final candidate varies based on the ultimate objective; for instance, prioritizing accuracy in coding tasks versus commonsense reasoning tasks. 

All distillation runs use FP8 precision, a softmax temperature of 1.0, which controls the softness of the probability distribution during knowledge distillation~\citep{hinton2015distilling}, sequence length of 8192, and a batch size of 768 samples.

\begin{table}[t]\small
\centering
\begin{tabular}{@{}lcccc@{}}
\toprule
  \textbf{Benchmark} &
  \textbf{\begin{tabular}[c]{@{}c@{}}Nemotron-H \\ 56B-Base\end{tabular}} &
  \textbf{\begin{tabular}[c]{@{}c@{}}Candidate 1\\ 47B-pruned\end{tabular}} &
  \textbf{\begin{tabular}[c]{@{}c@{}}Candidate 2\\ 47B-pruned\end{tabular}} &
  \textbf{\begin{tabular}[c]{@{}c@{}}Candidate 3\\ 47B-pruned\end{tabular}} \\ \toprule

MMLU & 84.2 & \textbf{81.5} $\rightarrow$ \textbf{83.7} & 80.8 $\rightarrow$ 83.2 & 81.3 $\rightarrow$ 83.6\\
Commonsense understanding (average) & 67.6 & 64.2 $\rightarrow$ 66.6 & 63.8 $\rightarrow$ \textbf{67.2} & \textbf{64.5} $\rightarrow$ 66.8\\
GSM8k & 91.4 & \textbf{67.9} $\rightarrow$ 89.2 & 60.7 $\rightarrow$ \textbf{91.1} & 19.5 $\rightarrow$ 90.5\\
Code (average) & 67.0 & 52.0 $\rightarrow$ 64.9 & \textbf{52.7} $\rightarrow$ 64.6 & 23.0 $\rightarrow$ \textbf{64.9}\\
\midrule
AVERAGE  & 70.6 & \textbf{63.3} $\rightarrow$ 69.3 & 62.1 $\rightarrow$ \textbf{69.6} & 53.7 $\rightarrow$ 69.5\\
\bottomrule
\end{tabular}
\caption{Comparison of benchmark scores for pruned candidates before and after lightweight distillation. Scores might be slightly different to those presented in Table~\ref{tab:accuracy_comparison_56b} due to minor differences in evaluation settings and averages over different sets of tasks.}
\label{tab:zero_shot_scores}
\end{table}

\subsection{Results}
Using the MiniPuzzle approach, we efficiently compress the 56B model to 47B parameters by pruning full layers (depth) and FFN hidden dimension size, improving inference speed and enabling longer-context inference on a 32GiB NVIDIA GeForce RTX 5090 GPU. The 47B model retains half the self-attention layers of 56B (5 instead of 10), along with 44 Mamba-2 layers (down from 54) and 49 FFN layers (down from 54). Additionally, the FFN intermediate size was pruned from 32768 to 30720. 
As shown in Figure~\ref{fig:inference_speed_56b} and Table~\ref{tab:accuracy_comparison_56b}, the resulting 47B model achieves a 1.2$\times$ faster inference on long contexts and near-lossless accuracy on benchmarks while requiring roughly $300\times$ fewer tokens compared to training from scratch.

\section{Vision-Language Models}
\label{sec:vlm}

In this section, we introduce Nemotron-H-8B-VLM and Nemotron-H-56B-VLM, which are built on Nemotron-H-8B-Instruct and Nemotron-H-56B-Base.
Vision-Language Models (VLMs) are generally recommended to be built on aligned models to enhance instruction-following capabilities~\citep{dai2024nvlm}.
In this work, we chose to build Nemotron-H-8B-VLM and Nemotron-H-56B-VLM on Nemotron-H-8B-Instruct and Nemotron-H-56B-Base (since Nemotron-H-56B-Instruct was unavailable).
Both Nemotron-H-8B-VLM and Nemotron-H-56B-VLM have been used as the base models for building Cosmos-Reason1~\citep{nvidia2025cosmosreason1physicalcommonsense}, a family of frontier reasoning models for physical AI.

\subsection{Model Architecture}
In previous studies, VLMs were constructed by integrating a text-only LLM with a vision encoder. Two prevalent architectures for this integration are the decoder-only architecture~\citep{liu2023visual} and the cross-attention-based architecture~\citep{alayrac2022flamingo}.
We use the decoder-only NVLM-D architecture~\citep{dai2024nvlm} due to its simplicity, parameter efficiency, ability to process high-resolution image inputs, and unified handling of multiple modalities by mapping non-text tokens (e.g., images, audio, videos) into the same embedding space as text tokens.
Nemotron-H-VLM comprises a vision encoder, a projector (a two-layer FFN), and the Nemotron-H LLM backbone. 

\paragraph{Vision encoder.}
For both the 8B and 56B VLMs, we use InternViT-300M-V2.5~\citep{chen2024expanding} as the vision encoder. It processes static 448$\times$448 pixel images as input with patch size 14$\times$14, and generates 1024 visual tokens in total. Each visual token is represented by a 1024-dimensional vector. 
Following the NVLM design~\citep{dai2024nvlm}, we dynamically resize the input image to the closest predefined aspect ratio $\{a~:~b\}$ ($a$ and $b$ are integers) based on its resolution and segment it into $a\times b\le$~12 tiles\footnote{The predefined ratios are $\{a~:~b\} = \{1:1,~ 1:2,~ 1:3,~ 1:4,~ 1:5,~ 1:6,~ 1:7,~ 1:8,~ 1:9,~ 1:10,~ 1:11,~ 1:2,~ 2:1,~ 2:2,~ 2:3,~ 2:4,~ 2:5,~ 2:6,~ 3:1,~ 3:2,~ 3:3,~ 3:4,~ 4:1,~ 4:2,~ 4:3,~ 5:1,~ 5:2,~ 6:1,~ 6:2,~ 7:1,~ 8:1,~ 9:1,~ 10:1,~ 11:1,~ 12:1\}$.}, each corresponding to a 448$\times$448-pixel image tile.
To preserve global context, we also generate a thumbnail 448$\times$448 tile, which is a scaled-down version of the whole image.
For example, given a 1920×1080 image, the closest predefined aspect ratio is $3:2$. The image is resized to 1344×896 and divided into $3\times2+1=7$ tiles, including one thumbnail tile.

To reduce processing overhead for the LLM, we downsample 1024 tokens to 256 by grouping four neighboring image tokens and concatenating them along the channel dimension. 
The image tokens from multiple tiles are concatenated with an interleaved tile ID tag in raw text format, which gives the downstream LLM information about the dynamic tiling structure; we find this is crucial to improving accuracies on various vision-language benchmarks.
For more details, see the NVLM technical report~\citep{dai2024nvlm}.
The concatenated visual tokens are processed by a two-layer FFN block, which maps each visual token into the text token embedding space. These embeddings are then fed into the LLM backbone.

\subsection{Training Method and Data}
Following NVLM~\citep{dai2024nvlm}, Nemotron-H-VLM is trained in two stages:
\begin{enumerate}
    \item \textbf{VL pre-training.} We train only the two-layer FFN for modality alignment while keeping both the Nemotron-H backbone and vision encoder frozen.
    \item \textbf{VL SFT.}  We fine-tune the vision encoder, FFN projector, and Nemotron-H backbone end-to-end on various task-oriented SFT data.
\end{enumerate}

For VL pre-training, we utilize a large and diverse image-text pre-training dataset from NVLM~\citep{dai2024nvlm}, including captioning~\citep{lin2014microsoft, sharma2018conceptual, ordonez2011im2text, li2022blip}, visual question answering~(VQA) on natural image~\citep{goyal2017making, krishna2017visual}, visual chart~\citep{kafle2018dvqa} and document understanding~\citep{Docmatix2024},  optical character recognition~(OCR)~\citep{mishra2019ocr} and scene-text recognition~\citep{veit2016coco}, and visual math reasoning~\citep{lindstrom2022clevr} data.
Overall, our VL pre-training dataset consists of 130 million samples.
We apply careful sanitization to remove any potentially harmful content.

For VL SFT, we leverage diverse and high-quality datasets from NVLM~\citep{dai2024nvlm} and Eagle2~\citep{li2025eagle}.
In addition to the previously mentioned categories, the dataset also includes knowledge-based VQA~\citep{marino2019ok}, visual reasoning~\citep{hudson2019gqa}, science-related VQA~\citep{lu2022learn}, and visual instruction-following data.  Overall, our VL SFT dataset consists of 6 million image-text samples. \citet{dai2024nvlm} and \citet{li2025eagle} have more details.

\begin{table}[t!]\small
\centering
\begin{tabular}{@{}lccc@{}}
\toprule
\textbf{Task} &
  \textbf{\begin{tabular}[c]{@{}c@{}}Nemotron-H \\ 8B-VLM\end{tabular}} &
  \textbf{\begin{tabular}[c]{@{}c@{}}VLM w/ Llama3.1\\ 8B-Instruct\end{tabular}} &
  \textbf{\begin{tabular}[c]{@{}c@{}}VLM w/ Qwen2.5\\ 7B-Instruct\end{tabular}}  \\ \toprule
MMMU~(val)  & \textbf{51.3}  & 44.8 & 51.1 \\
MathVista  & 62.5  & 61.7 & \textbf{63.8} \\
ChartQA        &  84.8	 & \textbf{85.4}  & 84.9 \\
AI2D          & 89.6  &  90.9 & \textbf{91.0} \\
OCRBench         & 840 & \textbf{847} & 836 \\
TextVQA           & 76.2  & \textbf{76.4} & 76.1 \\
RealWorldQA      & \textbf{62.2} & 60.9 & 60.6 \\
DocVQA            & 90.6  & \textbf{91.3} & 91.2 \\
\bottomrule
\end{tabular}
\caption{Evaluation of Nemotron-H-8B-VLM on vision-language benchmarks. We compare to the VLMs built with Llama-3.1-8B-Instruct and Qwen2.5-7B-Instruct using the same training recipe.}
\label{tab:nemotron-h-8b-vlm}
\end{table}

\begin{table}[t!]\small
\centering
\begin{tabular}{@{}lccc@{}}
\toprule
\textbf{Task} &
  \textbf{\begin{tabular}[c]{@{}c@{}}Nemotron-H \\ 56B-VLM\end{tabular}} &
  \textbf{\begin{tabular}[c]{@{}c@{}}VLM w/ Qwen2.5\\ 72B-Instruct\end{tabular}} &
  \textbf{\begin{tabular}[c]{@{}c@{}} NVLM-D-1.0 \\ 72B \normalfont{\footnotesize(2024-09-17)}\end{tabular}}  \\ \toprule
MMMU~(val)  & 63.6  & \textbf{65.1} & 62.6$^\text{†}$ \\
MathVista  & \textbf{70.7}  & 70.5 & 66.7$^\text{†}$  \\
ChartQA        & \textbf{89.4}	 & 88.9  & 86.0 \\
AI2D          & 94.7  &  \textbf{94.9} & 94.2 \\
OCRBench         & 862 & \textbf{869} & 853 \\
TextVQA           & 81.1  & \textbf{83.5} & 82.1 \\
RealWorldQA      & 68.4 & \textbf{71.4} & 69.7 \\
DocVQA            & \textbf{93.2}  & 92.0 & 92.6 \\
\bottomrule
\end{tabular}
\caption{We evaluate Nemotron-H-56B-VLM on a range of vision-language benchmarks, and compare it to a VLM built using Qwen2.5-72B-Instruct with the same training recipe. We also compare with NVLM-1.0-D 72B, a previous state-of-the-art VLM developed by NVIDIA. $^\text{†}$We evaluate NVLM-1.0-D 72B on MMMU and MathVista using VLMEvalKit~\citep{duan2024vlmevalkit}; this is the same evaluation setup used for other models and yields better results than the official numbers.}
\label{tab:nemotron-h-56b-vlm}
\end{table}

\subsection{Vision-Language Benchmark Results}
We evaluate Nemotron-H-8B-VLM and Nemotron-H-56B-VLM on a comprehensive set of vision-language benchmarks, including MMMU~\citep{yue2023mmmu}, MathVista~\citep{lu2024mathvista}, ChartQA \citep{masry2022chartqa}, AI2D~\citep{kembhavi2016diagram}, OCRBench~\citep{liu2024OCRBench}, TextVQA~\citep{singh2019towards}, RealWorldQA~\citep{xai2024Grok1-5}, and DocVQA~\citep{mathew2021docvqa}. These benchmarks assess a broad range of capabilities such as multimodal reasoning, mathematical reasoning in visual contexts, natural image understanding, scene-text recognition, chart \& document understanding, real-world perception, and OCR.

We compare Nemotron-H-8B-VLM with VLMs built on LLaMA-3.1-8B-Instruct and Qwen2.5-7B-Instruct, using the same training methodology and dataset. As shown in Table~\ref{tab:nemotron-h-8b-vlm}, Nemotron-H-8B proves to be a strong LLM backbone for developing best-in-class VLMs.
We further compare Nemotron-H-56B-VLM with NVLM-1.0-D~\citep{dai2024nvlm} and the VLM built with Qwen2.5-72B-Instruct using the same training recipe in Table~\ref{tab:nemotron-h-56b-vlm}.
Nemotron-H-56B-VLM achieves state-of-the-art results, demonstrating superior quality compared to previous models.

\section{Nemotron-H Reasoning Models}
\label{sec:alignment}
\subsection{Nemotron-H Vs. Transformer Model Alignment}
We first study the extent to which our hybrid Mamba-Transformer models can be effectively post-trained into instruction-tuned and long-context variants.
We conducted experiments on Transformer (Nemotron-T-8B-Exp-Base) and hybrid (Nemotron-H-8B-Exp-Base) models pretrained on identical data resulting in two models with very similar pretraining performance as seen in Table~\ref{tab:base_exp_model_comparison}.

\begin{table}[h!]\small
\centering
\setlength{\tabcolsep}{4pt}
\begin{tabular}{@{}lcc@{}}
\toprule
\textbf{Task} &
  \textbf{\begin{tabular}[c]{@{}c@{}}Nemotron-T\\ 8B-Exp-Base\end{tabular}} &
  \textbf{\begin{tabular}[c]{@{}c@{}}Nemotron-H\\ 8B-Exp-Base\end{tabular}} \\
\toprule
MMLU (5-shot) & \textbf{70.3} & 69.9 \\
GSM8k & \textbf{64.8} & 64.1 \\
MBPP+ (0-shot greedy pass@1) & 53.7 & \textbf{54.5} \\
PIQA (0-shot) & 81.1 & \textbf{82.8} \\
OpenBookQA (0-shot) & 45.4 & \textbf{46.2} \\
\bottomrule
\end{tabular}
\caption{Comparison of Transformer and hybrid base models on selected tasks.}
\label{tab:base_exp_model_comparison}
\end{table}

We then post-trained both base models into instruction-tuned variants, Nemotron-T-8B-Exp-Instruct and Nemotron-H-8B-Exp-Instruct using a multistage training procedure. In \texttt{stage1}, we perform supervised fine-tuning (SFT) on a blend of 6 million samples including code, math, and general instruction-following tasks. To improve long-context performance, we include training on extended conversations up to 512k tokens by concatenating shorter samples. We further enhance context by inserting references to earlier turns and introducing semantically related segments across the conversation. This improves RULER~\citep{hsieh2024ruler} scores even at shorter context lengths (e.g., 128k).

In \texttt{stage2}, we switch to preference tuning using offline RPO~\citep{sun2025rewardawarepreferenceoptimizationunified} on general-domain prompts. We then perform an additional RPO round focused on narrow instruction following, following IFEval~\citep{zhou2023instruction} style prompts. Samples are randomly extended to 32k tokens using conversations from \texttt{stage1}, preserving long-context capabilities.

In \texttt{stage3}, we apply on-policy RPO, generating both preferred and dispreferred completions using the \texttt{stage2} checkpoint. We incorporate safety data from AEGIS2.0~\citep{ghosh2025aegis2} and increase reward scaling, which improves performance on downstream benchmarks.

We experimented with multiple rounds of iterative offline RPO and DPO. Both yielded similar outcomes; we report results for the best model from either method. The best Transformer model used three RPO rounds; the best hybrid model used two DPO rounds.

Table~\ref{tab:instruct_exp_model_comparison} shows the results. These findings support that hybrid Mamba-Transformer models can be post-trained to match or exceed Transformer performance.
\begin{table}[t!]\small
\centering
\setlength{\tabcolsep}{4pt}
\begin{tabular}{@{}lcc@{}}
\toprule
\textbf{Benchmark} &
  \textbf{\begin{tabular}[c]{@{}c@{}}Nemotron-T\\ 8B-Exp-Instruct\end{tabular}} &
  \textbf{\begin{tabular}[c]{@{}c@{}}Nemotron-H\\ 8B-Exp-Instruct\end{tabular}} \\
\toprule
MT-Bench & 7.9 & \textbf{8.0} \\
MMLU (0-shot generative) & \textbf{68.0} & 67.4 \\
MBPP+ & 65.6 & \textbf{66.9} \\
MATH & \textbf{68.8} & 68.5 \\
IFEval Prompt & 74.3 & \textbf{74.7} \\
IFEval Instruct & 81.8 & \textbf{82.1} \\
BFCL (V2) & 64.3 & \textbf{65.4} \\
\bottomrule
\end{tabular}
\caption{Comparison of instruction-tuned Transformer and hybrid models on alignment benchmarks.}
\label{tab:instruct_exp_model_comparison}
\end{table}

\subsection{Training Nemotron-H Reasoning Models}
Following the pilot study in which we found that mamba2-hybrid models and pure transformer models have similar post-training properties we proceed to train our Nemotron-H 8B Reasoning and Nemotron-H 47B Reasoning models.

\subsubsection{Supervised Finetuning Stages}
The training pipeline began with supervised fine-tuning (SFT), using curated examples that include explicit reasoning traces—enclosed in \verb|<think>...</think>| tags—to guide the model through step-by-step problem-solving before reaching a final answer. These traces often represent multiple possible solution paths and encourage the model to explore alternatives and iterate, improving accuracy. However, the added verbosity also increases inference cost, especially for longer traces.
To balance this, we introduced paired examples with the reasoning stripped out, allowing the model to learn when and how to respond directly. 
This dual-format training helps the model adapt fluidly to different reasoning requirements.

\paragraph{Stage 1: Mastering Math, Code and Science Reasoning}
The first phase of fine-tuning focused on math, science, and coding—domains where explicit reasoning is especially valuable. The training data here used a 5:1 ratio of reasoning to non-reasoning samples, and some of these examples are available publicly in the Llama-Nemotron-Post-Training-Dataset. 

\paragraph{Stage 2: Expanding Instructional Coverage, Dialogue and Safety}
The second phase shifted toward instruction following, safety alignment, and multi-turn dialogue—while continuing to sample from Stage 1 to retain strong STEM performance. This dataset was more compact—about 10× smaller—and offered a balanced mix of reasoning and non-reasoning samples. This helped the model generalize across a broader range of tasks while improving control over reasoning mode switching.

\subsubsection{Long-Context Training}
\begin{figure*}[!t]       
  \begin{minipage}[t]{0.48\textwidth}
    \centering
    \includegraphics[width=\linewidth]{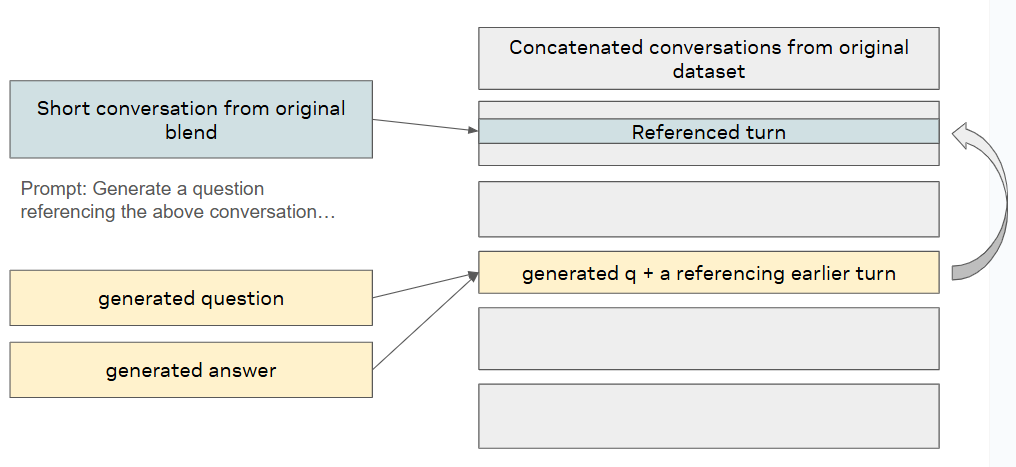}
    \captionof{figure}{After selecting a short conversation from the original blend, we generate multiple turns referencing turns from that conversation. These turns are inserted at some point later in the extended sample.}
    \label{fig:long_context_1}
  \end{minipage}\hfill
  \begin{minipage}[t]{0.48\textwidth}
    \centering
    \includegraphics[width=\linewidth]{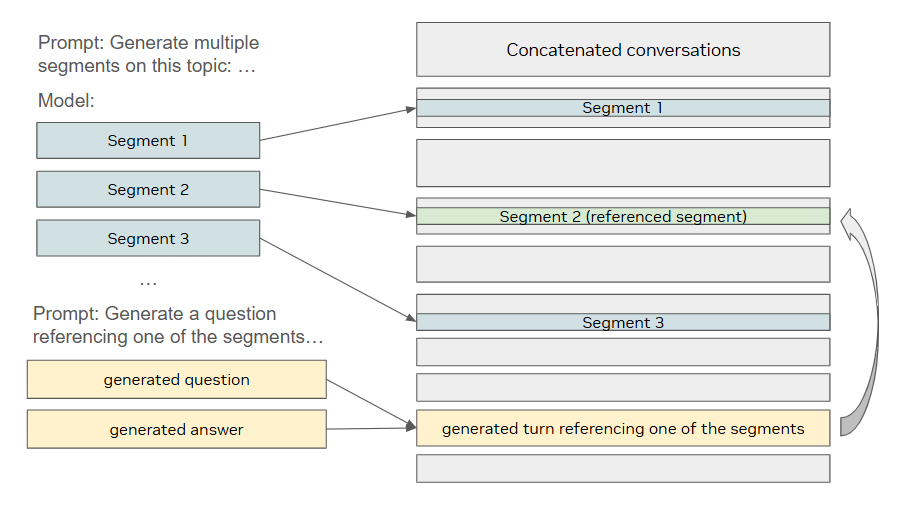}
    \captionof{figure}{Multiple segments are generated with a unified theme and scattered throughout the context. Afterwards, we generate turns that that refer to one or more of these segments which are placed later in the sample.}
    \label{fig:long_context_2}
  \end{minipage}
\end{figure*}

In order to improve long-context capabilities, we do additional training on an augmented version of our Stage 2 blend, consisting of samples up to 512k tokens each. The samples in this extended blend are created by concatenating the original Stage 2 SFT conversations, as well as adding several additional types of turns, with the goal of improving the long-range memory of the model. The added turns are constructed as follows:
 
\begin{enumerate}[leftmargin=*]
      \item \textbf{Long-range references to existing conversations.}  
        A random short conversation that appears earlier in the concatenated sample is selected; we then generate follow-up user queries that explicitly references this turn sequence by topic (e.g., “In the part of our conversation where we discussed modern authors...”).  
        Both the referencing questions and answers are generated with DeepSeek-R1 \citep{deepseek}
        These reference pairs are inserted throughout conversations in both the reasoning and non-reasoning subsets of the data blend (see Figure~\ref{fig:long_context_1}). This method does not require relying on the generator model's long context understanding, as we only need the referenced conversation as context.

      \item \textbf{Multi-turn references}
        In addition to turns referencing conversations from the short blend, we also generate multiple turns with a unified topic that may reference each other. Topics for these sequences are chosen as a combination of 3 random keywords. We then generate multiple follow-up turns that each reference one or more of these segments, requiring the model to determine which of these related segments is being referred to. All of these turns are then inserted randomly thoughout the extended conversations (see Figure~\ref{fig:long_context_2}). Similarly to the previous type, this does not require a long-context generator.

      \item \textbf{Document aggregation QA.}  
        A user turn embeds a long document (up to 32k tokens) drawn from the pre-training data. The turn is followed by several questions whose answers require integrating information scattered across the long document. 

      \item \textbf{Common-word retrieval with distractors.}  
        To target the RULER “common-words” subtask, we insert user turns that list 30-80k tokens comprising a list of keywords interleaved with random distractors.  
        The assistant is asked to return the $n$ most frequent  words.
        This specialised task is used only in the non-CoT subset. It is created without the use of a language model.
\end{enumerate}

 We find that hybrid models can learn to operate on higher context length within just a few hundred steps of training on these 512k-token extended blends. We evaluated this capability using  the RULER~\citep{hsieh2024ruler} benchmark in non-reasoning mode, with the model achieving an 83\% RULER score on 128k sequence length after this stage.

\subsubsection{Reinforcement Learning with GRPO}
After SFT, we applied Group Relative Policy Optimization (GRPO) in multiple phases~\citep{shao2024deepseekmath}. Each phase targeted specific skills—like instruction following or tool use—by creating task-specific datasets with automatic verifiers, followed by broader fine-tuning with a general reward model.

\paragraph{Instruction-Following Tuning} To strengthen instruction adherence, we sampled 16,000 prompts from the LMSYS Chat dataset and paired them with IFEval-style instructions. A rule-based verifier scored outputs based on how well they satisfied each instruction, creating a reward signal that prioritized following directions with precision. We have used TRT-LLM backend for policy rollouts. IFEval RL experiments provided significant boost to IFEval capabilities while the rest of the benchmarks fluctuated slightly requiring careful checkpoint selection. We have used a learning rate of $1\mathrm{e}{-6}$ which showed consistent improvement in instruction following capabilities without regressions in other model capabilities.

\paragraph{General Helpfulness via Reward Model} In the final RL phase, we introduced a Qwen-32B-based reward model (scoring 92.8 on RewardBench) to improve overall response helpfulness. We trained using GRPO and prompts from HelpSteer2~\cite{Wang2024HelpSteer2}. This final stage led to noticeable gains in output quality.

\subsubsection{Controlled Reasoning Inference}
Inference-time behavior can be customized using simple control tags in the system prompt:

\begin{itemize}
\item \verb|{`reasoning': True}| triggers reasoning mode
\item \verb|{`reasoning': False}| triggers direct-answer mode
\item Omitting the tag lets the model choose
\end{itemize}

Our Jinja chat template detects these control strings and modifies the assistant’s response accordingly. When \verb|{`reasoning': True}| is present, the response is prefixed with \verb|Assistant:<think>\n|, indicating the start of a reasoning trace. When \verb|{`reasoning': False}| is found, the response is prefixed with \verb|Assistant:<think></think>|, signaling a non-reasoning response. This mechanism resulted in near 100\% control of reasoning or non-reasoning modes.

\subsection{Results}

\subsubsection{Nemotron-H-47B-Reasoning-128K}
Across benchmarks in math, coding, science, tool use, and dialogue, \textbf{Nemotron-H-47B-Reasoning-128K} achieves accuracy on par with or better than Llama-Nemotron Super 49B V1.0, and outperforms Qwen3 32B on all non-coding benchmarks. The model supports post-training quantization of all linear layers, enabling efficient deployment with minimal accuracy loss. We also provide the quantized checkpoint and its corresponding results to demonstrate effectiveness in practical settings.

\begin{table}[t!]\small
\centering
\setlength{\tabcolsep}{4pt}
\begin{tabular}{@{}lcccc@{}}
\toprule
\textbf{Benchmark} &
  \textbf{\begin{tabular}[c]{@{}c@{}}Llama-Nemotron\\ Super V1\end{tabular}} &
  \textbf{\begin{tabular}[c]{@{}c@{}}Nemotron-H 47B\\ Reasoning\end{tabular}} &
  \textbf{\begin{tabular}[c]{@{}c@{}}Nemotron-H 47B\\ Reasoning (FP8)\end{tabular}} &
  \textbf{\begin{tabular}[c]{@{}c@{}}Qwen3\\ 32B\end{tabular}} \\
\toprule
AIME25              & 53.0  & 54.2  & 54.2  & \textbf{55.8} \\
MATH500             & 96.6  & 96.2  & \textbf{96.8} & 91.4 \\
GPQA-D              & 65.1  & \textbf{65.7} & 64.6  & 65.2 \\
MBPP                & 90.2  & 91.8  & 91.9  & \textbf{95.5} \\
MBPP\_PLUS          & 75.1  & 79.9  & 79.9  & \textbf{81.5} \\
LCB                 & 41.2  & 50.2  & 53.4  & \textbf{64.2} \\
BFCL                & 72.7  & 72.9  & \textbf{73.2} & 71.3 \\
IFEVAL-Prompt       & 81.8  & \textbf{84.5} & 84.5  & 83.9 \\
IFEVAL-Instruction  & 87.4  & \textbf{89.7} & 89.6  & 88.6 \\
Arena Hard          & 85.0  & 85.0  & 80.5       & \textbf{93.8} \\
\bottomrule
\end{tabular}
\caption{Comparison of benchmark performance across multiple model variants. Best scores per row are shown in bold.}
\label{tab:model_comparison}
\end{table}

\subsubsection{Nemotron-H-8B-Reasoning-128K}

\paragraph{Checkpoint merging}
We observed that the smaller 8B model struggled on  internal safety benchmarks. To address this, we opted for checkpoint interpolation~\cite{wortsman2022modelsoup}, blending in an earlier RL checkpoint aligned from a safer long-context extended checkpoint. Checkpoint interpolation is performed by linearly interpolating model weights: $(1-\alpha) * w_{model1} + \alpha * w_{model2}$. We experimented with a parameter sweep over $\alpha$ values from 0.1 to 0.9 in increments of 0.1, and found that values around 0.5 offered a good trade-off between reasoning capabilities and safe behavior.

Across benchmarks in AIME25, science, tool use, and instruction-following, \textbf{Nemotron-H-8B-Reasoning-128K} achieves accuracy on par with or better than Llama-Nemotron Nano 8B V1.0. The model supports post-training quantization of all linear layers, enabling efficient deployment with minimal accuracy loss. We also provide the FP8 quantized checkpoint and its corresponding results to demonstrate effectiveness in practical settings.

\begin{table}[t!]\small
\centering
\setlength{\tabcolsep}{4pt}
\begin{tabular}{@{}lcccc@{}}
\toprule
\textbf{Benchmark} &
  \textbf{\begin{tabular}[c]{@{}c@{}}Llama-Nemotron\\ Nano 8B v1.0\end{tabular}} &
  \textbf{\begin{tabular}[c]{@{}c@{}}Nemotron-H 8B\\ Reasoning\end{tabular}} &
  \textbf{\begin{tabular}[c]{@{}c@{}}Nemotron-H 8B\\ Reasoning (FP8)\end{tabular}}  \\
\toprule
AIME25              & 51.4  & \textbf{51.7}  & 46.7  \\
MATH500             & \textbf{94.8}  & 94.0  & 94.0  \\
GPQA-D              & 53.5  & \textbf{55.1}  & 54.5  \\
MBPP                & 84.1  & 86.0  & \textbf{86.2}  \\
MBPP\_PLUS          & 71.4  & \textbf{74.1}  & 73.3  \\
LCB                 & \textbf{52.6}  & 49.5  & 44.8  \\
BFCL                & 63.6  & 68.8  & \textbf{70.6}  \\
IFEVAL-Prompt       & \textbf{71.8}  & 71.4  & 71.5  \\
IFEVAL-Instruction  & 79.3  & 79.6  & \textbf{79.9}  \\
\bottomrule
\end{tabular}
\caption{Comparison of benchmark performance across multiple 8B model variants. Best scores per row are bolded.}
\label{tab:model_comparison_8b}
\end{table}
\section{Conclusions}
\label{sec:conclusions}
The Nemotron-H family of models demonstrates that hybrid model architectures can offer the best of both worlds: comparable to state-of-the-art Transformer models in terms of capabilities while offering improved inference speed. We find the Nemotron-H base models offer either better or on-par accuracy compared to other similarly-sized state-of-the-art open-sourced Transformer models (e.g., Qwen-2.5-7B/72B and Llama-3.1-8B/70B), while being up to 3$\times$ faster at inference. We also find these base models can be adapted to yield effective VLMs as well as instruction-tuned models. Additionally, both our FP8-based training recipe and MiniPuzzle can be used to reduce the cost of creating these models.

We have released the Nemotron-H base checkpoints described in this paper with support in Hugging Face and NeMo to facilitate further research:
\begin{itemize}
    \item \textbf{Nemotron-H-56B-Base.} \href{https://huggingface.co/nvidia/Nemotron-H-56B-Base-8K}{Hugging Face} and \href{https://catalog.ngc.nvidia.com/orgs/nvidia/teams/nemo/models/nemotron-h-56b-base-8k}{NGC}.
    \item \textbf{Nemotron-H-47B-Base.} \href{https://huggingface.co/nvidia/Nemotron-H-47B-Base-8K}{Hugging Face} and \href{https://catalog.ngc.nvidia.com/orgs/nvidia/teams/nemo/models/nemotron-h-47b-base-8k}{NGC}.
    \item \textbf{Nemotron-H-8B-Base.} \href{https://huggingface.co/nvidia/Nemotron-H-8B-Base-8K}{Hugging Face} and \href{https://catalog.ngc.nvidia.com/orgs/nvidia/teams/nemo/models/nemotron-h-8b-base-8k}{NGC}.
\end{itemize}

Nemotron-H Reasoning checkpoints described are also available in Hugging Face:
\begin{itemize}
    \item \textbf{Nemotron-H-47B-Reasoning.} \href{https://huggingface.co/nvidia/Nemotron-H-47B-Reasoning-128K}{Hugging Face}.
    \item \textbf{Nemotron-H-8B-Reasoning.} \href{https://huggingface.co/nvidia/Nemotron-H-8B-Reasoning-128K}{Hugging Face}.
\end{itemize}

\section*{Contributors}

We thank the following people for their invaluable contributions to Nemotron-H.

\textbf{Data.} Abhinav Khattar, Aleksander Ficek, Amala Sanjay Deshmukh, Andrew Tao, Ayush Dattagupta, Brandon Norick*, Chengyu Dong*, Dan Su*, Daria Gitman, Evelina Bakhturina, Igor Gitman, Ilia Karmanov, Ivan Moshkov, Jane Polak Scowcroft, Jarno Seppanen, Jiaxuan You, Jocelyn Huang, John Kamalu*, Joseph Jennings*, Jupinder Parmar*, Karan Sapra, Kateryna Chumachenko, Kezhi Kong*, Lukas Voegtle, Lindsey Pavao, Markus Kliegl*, Matvei Novikov, Mehrzad Samadi, Miguel Martinez, Mostofa Patwary*, Osvald Nitski, Philipp Fischer, Pritam Gundecha, Rabeeh Karimi Mahabadi, Sean Narenthiran, Sanjeev Satheesh*, Seungju Han, Shrimai Prabhumoye*, Shubham Pachori, Shubham Toshniwal, Siddhartha Jain, Somshubra Majumdar, Syeda Nahida Akter*, Timo Roman, Ushnish De, Vahid Noroozi, Vitaly Kurin, Wasi Uddin Ahmad, Wei Du, Yao Xu, Yejin Choi, Ying Lin*.

\textbf{FP8 recipe.} Carlo del Mundo, Dusan Stosic, Eric Chung, Jinze Xue, John Kamalu, Kirthi Sivamani, Mike Chrzanowski*, Mohammad Shoeybi, Mostofa Patwary, Oleg Rybakov*, Paulius Micikevicius, Peter Dykas*, Przemek Tredak, Zhongbo Zhu.

\textbf{Model architecture.} Brandon Norick*, Duncan Riach*, Roger Waleffe*, Wonmin Byeon*.

\textbf{Pre-training.} Deepak Narayanan*, Hongbin Liu, Kunlun Li, Maciej Bala, Michael Andersch, Mikolaj Blaz, Oleg Rybakov, Peter Dykas, Roger Waleffe*, Sangkug Lym, Selvaraj Anandaraj, Seonmyeong Bak, Slawek Kierat, Szymon Migacz, Xiaowei Ren.

\textbf{Infrastructure.} Aaron Blakeman, Aarti Basant, Ashwin Poojary, Brian Butterfield, Christine Harvey, Ding Ma, Dong Ahn, Gargi Prasad, Hui Li, Jason Sewall, Jing Zhang, Jining Huang, Kumar Anik, Maer Rodrigues de Melo, Mohamed Fawzy, Ning Xu, Pasha Shamis, Pierre-Yves Aquilanti, Rahul Kandu, Ruoxi Zhang, Sabrina Kavanaugh, Sergey Kashirsky, Shelby Thomas, Sirshak Das, Sriharsha Niverty, Stefania Alborghetti, Tal Shiri.

\textbf{Distillation.} Akhiad Bercovich*, Ali Taghibakhshi*, Daniel Korzekwa, Elad Segal*, Izik Golan*, Marcin Chochowski*, Mostofa Patwary, Pavlo Molchanov*, Ran El-Yaniv, Raviraj Joshi, Roger Waleffe, Saurav Muralidharan*, Sharath Turuvekere Sreenivas*, Tomer Ronen*.

\textbf{VLM.} Andrew Tao, Boxin Wang*, Danny Yin, Fuxiao Liu, Guilin Liu, Guo Chen, Jason Lu, Jon Barker, Lukas Voegtle, Matthieu Le, Mike Ranzinger, Nayeon Lee*, Philipp Fischer, Song Han, Tuomas Rintamaki, Tyler Poon, Wei Ping*, Wenliang Dai*, Zhiding Yu, Zhiqi Li, Zhuolin Yang*.

\textbf{Alignment and long context.}  
Adithya Renduchintala*, Ali Taghibakhshi, Ameya Sunil Mahabaleshwarkar*,  
Bilal Kartal*, David Mosallanezhad, Dima Rekesh*, Ellie Evans,  
Fei Jia*, Felipe Sores*, Gerald Shen*, Haifeng Qian*,  
Hoo Chang Shin, Jiaqi Zeng, Julien Veron Vialard*, Luis Vega*,  
Makesh Narsimhan Sreedhar, Michael Evans, Olivier Delalleau,  
Prasoon Varshney, Samuel Kriman*, Shantanu Acharya*,  
Soumye Singhal, Tugrul Konuk,  
Yian Zhang*, Yoshi Suhara, Zijia Chen.

\textbf{Inference.} Helen Ngo*, Keshav Santhanam*, Vijay Korthikanti*.

\textbf{Software support.}  
Adithya Renduchintala, Ali Taghibakhshi, Anna Shors,  
Ashwath Aithal, Balaram Buddharaju, Bobby Chen,  
Cyril Meurillon, David Mosallanezhad, Deepak Narayanan,  
Dmytro Pykhtar, Duncan Riach, Ellie Evans,  
Fei Jia, Felipe Sores, Gerald Shen,  
Helen Ngo, Jared Casper, Jimmy Zhang,  
Keshav Santhanam, Lawrence McAfee, Luis Vega,  
Michael Evans, Nima Tajbakhsh, Olivier Delalleau,  
Parth Chadha, Piotr Bialecki, Roger Waleffe,  
Sahil Jain, Terry Kong, Tyler Poon,  
Vijay Korthikanti, Yian Zhang, Yoshi Suhara, Zhiyu Li.

\textbf{Evaluations and safety.} Ameya Sunil Mahabaleshwarkar*, Christopher Parisien, David Mosallanezhad*, Denys Fridman, Eileen Long, Erick Galinkin, Ewa Dobrowolska, Katherine Luna, Leon Derczynski, Marta Stepniewska-Dziubinska, Michael Evans, Roger Waleffe*, Sanjeev Satheesh*, Shaona Ghosh, Shrimai Prabhumoye, Suseella Panguluri, Syeda Nahida Akter, Varun Singh.

\textbf{Program management.} Joey Conway, Krzysztof Pawelec, Shyamala Prayaga, Swetha Bhendigeri, Trisha Saar.

\textbf{Leadership.} Alexis Bjorlin, Andrew Tao*, Boris Ginsburg*, Bryan Catanzaro*, Eric Chung, Jan Kautz, Jonathan Cohen*, Kari Briski, Misha Smelyanskiy, Mohammad Shoeybi*, Mostofa Patwary*, Oleksii Kuchaiev*, Sharon Clay, Song Han, Timo Roman, Wei Ping*.

* Core contributor.

\newpage

\bibliography{references}
\bibliographystyle{references}

\end{document}